\documentclass{article}

\usepackage{microtype}
\usepackage{graphicx}
\usepackage{subfigure}
\usepackage{enumitem}
\usepackage{booktabs} % for professional tables

\usepackage{hyperref}

\usepackage[accepted]{icml2025}

\usepackage{amsmath}
\usepackage{amssymb}
\usepackage{mathtools}
\usepackage{amsthm}

\usepackage[utf8]{inputenc} % allow utf-8 input
\usepackage[T1]{fontenc}    % use 8-bit T1 fonts
\usepackage{url}            % simple URL typesetting
\usepackage{booktabs}       % professional-quality tables
\usepackage{amsfonts}       % blackboard math symbols
\usepackage{nicefrac}       % compact symbols for 1/2, etc.
\usepackage{microtype}      % microtypography
\usepackage{xcolor}         % colors

\usepackage{amssymb}
\usepackage{bbm}

\newcommand{\lei}[1]{{\color{black} #1}}

\usepackage{wrapfig}

\usepackage{booktabs}
\usepackage{graphicx}
\graphicspath{{assets/}} %Setting the graphicspath

\usepackage{lipsum}
\usepackage{makecell} % To make it easier to format the header
\usepackage{subfigure}

\usepackage{multirow}
\usepackage{enumitem}
\usepackage{colortbl}
\usepackage{subcaption}
\usepackage{hyperref} % [pagebackref,breaklinks,colorlinks]

\usepackage{pifont} % For check marks and crosses
\usepackage{xspace}
\makeatletter
\DeclareRobustCommand\onedot{\futurelet\@let@token\bmv@onedotaux}
\def\bmv@onedotaux{\ifx\@let@token.\else.\null\fi\xspace}
\makeatother
\def\eg{\emph{e.g}\onedot} 
\def\ie{\emph{i.e}\onedot} 
 
\def\etc{\emph{etc}\onedot} 
\def\wrt{w.r.t\onedot} 
\def\aka{a.k.a\onedot}

\makeatother

\usepackage{amsmath,amsfonts,bm}
\renewcommand\cdots{...}

\newcommand{\mX}{\mathbf{X}}

\newcommand{\vx}{\mathbf{x}}
\newcommand{\vq}{\mathbf{q}}

\newcommand{\mbr}[1]{\mathbb{R}^{#1}}

\newcommand{\vv}{\mathbf{v}}

\newcommand{\idx}[1]{\mathcal{I}_{#1}}

\newcommand{\vz}{\mathbf{z}}

\def\eg{\emph{e.g.}}

\newcommand{\stkout}[1]{{\ifmmode\text{\sout{\ensuremath{#1}}}\else\sout{#1}\fi}}

\usepackage[capitalize,noabbrev]{cleveref}

\theoremstyle{plain}

\theoremstyle{definition}

\theoremstyle{remark}

\usepackage[textsize=tiny]{todonotes}
\usepackage{caption}

\icmltitlerunning{Optimizing Calibration by Gaining Aware of
Prediction Correctness}

\begin{document}

\twocolumn[{
\icmltitle{Optimizing Calibration by Gaining Aware of
Prediction Correctness}



\icmlsetsymbol{equal}{*}

\begin{icmlauthorlist}

\icmlauthor{Yuchi Liu}{ANU}
\icmlauthor{Lei Wang}{ANU,d61}
\icmlauthor{Yuli Zou}{PolyU}
\icmlauthor{James Zou}{Stanford}
\icmlauthor{Liang Zheng}{ANU}
\end{icmlauthorlist}

\icmlaffiliation{ANU}{School of Computing, The Australian National University (ANU), Canberra, Australia}
\icmlaffiliation{d61}{Data61, The Commonwealth Scientific and Industrial Research Organisation (CSIRO), Canberra, Australia}
\icmlaffiliation{PolyU}{The Hong Kong Polytechnic University (PolyU), Hong Kong, China}
\icmlaffiliation{Stanford}{Department of Computer Science, Stanford University, California, United States}

\icmlcorrespondingauthor{Yuchi Liu}{yuchi.liu@anu.edu.au}

\icmlkeywords{Machine Learning, ICML}

\begin{center}
    \centering
    \includegraphics[width=\linewidth]{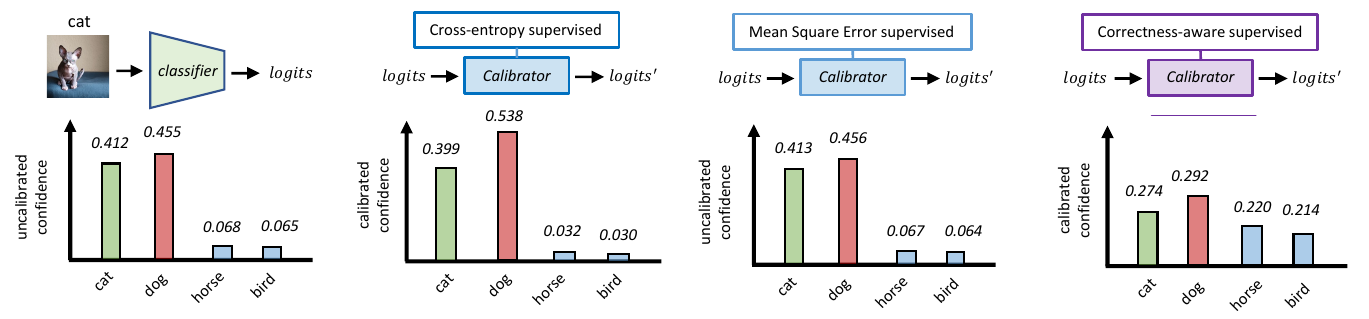}
    \vskip -0.1in
     \captionof{figure}{\textbf{A failure example for calibrators trained by the Cross-Entropy (CE) or Mean Square Error (MSE) loss.} % Different behaviors of two calibrators that trained by Cross-entropy loss and Correctness-aware loss, respectively.} 
    A classifier makes a wrong prediction of a cat image. 
    Before calibration, the classifier gives probabilities of 0.412 and 0.455 on the ground-truth and predicted classes, respectively. 
    The calibrator trained with the CE loss assigns even higher confidence 0.538 to the wrong class, making things worse, and that trained by MSE maintains a similar confidence 0.456. 
    %the calibrator learned from cross-entropy loss tend to maintain high confidence value on the ground truth class resulting a increase in prediction confidence (0.474 $\rightarrow$ 0.538). 
    In comparison, calibrator trained with the proposed Correctness-Aware (CA) loss effectively decreases confidence of this wrong prediction to 0.292, improving calibration. % Section~\ref{sec:exp} shows CE and MSE losses are limited for such narrowly wrong predictions (\eg, probability on ground truth is not far from confidence), often seen in OOD data.
    \label{fig:first_fig}
    }  
\end{center}

\vskip 0.3in
}]



\printAffiliationsAndNotice{\icmlEqualContribution} % otherwise use the standard text.

\begin{abstract}
Model calibration aims to align confidence with prediction correctness.
The Cross-Entropy (CE) loss is widely used for calibrator training, which enforces the model to increase confidence on the ground truth class. However, we find the CE loss has intrinsic limitations. For example, for a narrow misclassification (\eg, a test sample is wrongly classified and its softmax score on the ground truth class is 0.4), a calibrator trained by the CE loss often produces high confidence on the wrongly predicted class, which is undesirable. In this paper, we propose a new post-hoc calibration objective derived from the aim of calibration. Intuitively, the proposed objective function asks that the calibrator decrease model confidence on wrongly predicted samples and increase confidence on correctly predicted samples. Because a sample itself has insufficient ability to indicate correctness, we use its transformed versions (\eg, rotated, greyscaled, and color-jittered) during calibrator training. Trained on an in-distribution validation set and tested with isolated, individual test samples, our method achieves competitive calibration performance on both in-distribution and out-of-distribution test sets compared with the state of the art. Further, our analysis points out the difference between our method and commonly used objectives such as CE loss and Mean Square Error (MSE) loss, where the latters sometimes deviates from the calibration aim. 
\end{abstract}

\section{Introduction}
\label{sec:intro}

Model calibration is an important technique to enhance the reliability of machine learning systems. 
Generally, it aims to align predictive uncertainty (\aka confidence) with prediction accuracy. 
We are interested in post-hoc accuracy preserving calibrators that scale the model output to make it calibrated \citep{guo2017calibration, zhang2020mix, tomani2022parameterized, gupta2020calibration, kull2019beyond}.

Existing methods typically use Maximum Likelihood Estimation (MLE) to train a calibrator for the classification task, such as the Mean Square Error (MSE) loss \citep{tomani2022parameterized, tomani2023beyond, zhang2020mix} and the Cross-Entropy (CE) loss \citep{guo2017calibration, zou2023adaptive}. Although these approaches demonstrate efficacy in reducing calibration errors such as Expected Calibration Error (ECE) and Brier scores, they lack theoretical guarantee that the calibration error is minimized when MLE converges. In Fig.~\ref{fig:first_fig}, for an image which is incorrectly classified and has a relatively high probability on the ground truth class, calibrators trained by the CE or MSE loss would give high confidence on the wrongly predicted class. It means a user may trust this prediction to be true. As to be revealed in Sec.~\ref{sec:theory_comparison}, the inherent problem of CE and MSE loss functions limits them in calibrating such test cases.% , but our method effectively reduces the confidence. 

In this paper, we derive a concrete interpretation of the goal of calibration which is then directly translated into a novel loss function that aligns with the newly interpreted goal. 
Specifically, we start from the general definition of calibration and its error \citep{guo2017calibration}, and then, under a finite test set, represent the error in a discretized form. Minimizing this discretized error gives us a very interesting and intuitive calibration goal: A correct prediction should have possibly high confidence and a wrong prediction should have possibly low confidence. 
%
% This goal can naturally reduce the number of samples that falls into the region of mid-level confidence (\eg, 0.4 -- 0.6) where users are hard to make decisions (accept or reject model predictions).
% 
Theoretically, this optimization goal can naturally reduce the overlap of confidence values between correct and incorrect predictions. %, which will improve calibration performance. %Consequently, users would make fewer mistakes when making decisions about accepting or rejecting model predictions based on confidence.

%Interestingly, through experimental results we show that such an objective function can also implicitly optimise the mid-level confidence (see Fig. ~\ref{})

% To achieve this goal, this paper (i) provides a concrete interpretation of the calibration goal and (ii) designs a loss function that aligns model confidence with prediction correctness. Unlike existing MLE-based calibration approaches, which % either
% maximize the confidence of the ground truth,
% %or match confidence with correctness, 
% our method (i) provides a concrete interpretation of the calibration goal and (ii) designs a loss function specifically focusing on mid-level confidence, while maintaining the efficiency of the original objective functions.
%
%Following this go, 
%We directly use this interpretation for calibration optimization. Specifically, 
We translate this goal into a loss function that enforces high confidence (\ie, 1) for correctly classified samples, and low confidence (\ie, $\frac{1}{C}$, where $C$ is the number of classes) for wrongly classified ones, named Correctness-Aware (CA) loss. % which closely aligns with calibration objective. 
Nevertheless, it is non-trivial to identify classification correctness. We propose to use transformed versions of original images as the calibrator input: consistency between their prediction results suggests prediction correctness of the original sample. 

%We show that this strategy is very useful, especially on out-of-distribution (OOD) test sets, where not only state-of-the-art calibration performance but also separable for correct and incorrect samples are observed.

Our method allows a calibrator to be trained on the in-distribution validation set and directly applied to individual test samples during inference. % , which is consistent with practical usage. 
We demonstrate the effectiveness of the proposed strategy on various test sets. In both in-distribution (IND) and out-of-distribution (OOD) test sets, our method is clearly superior to uncalibrated models and competitive compared with state-of-the-art  calibrators. Moreover, our method demonstrates the potential to better separate correct and incorrect test samples using their calibrated confidence. Below we summarize the main contributions.

\begin{itemize}[leftmargin=*]
    \item Theoretically, we derive the concrete goal of model calibration that has a clear semantic meaning. This allows us to design a new calibration loss function: correctly classified samples should have high confidence, while incorrectly classified ones with low confidence. 

    \item To indicate prediction correctness, we use the softmax prediction scores of transformed versions of the original image as calibrator inputs. 
    
    \item Our method achieves competitive calibration performance on various IND and OOD datasets. %Interestingly, under ImageNet and iWilds setups, we observe better separability of correctly and incorrectly classified samples using the calibrated confidence. % and incorrectly We design a calibration loss function based on this goal. Using transformed versions of a sample as calibrator input, we demonstrate state-of-the-art calibration performance on OOD test sets. 
   % \item Our calibration enhances the separability of confidence levels between correct and incorrect samples. % Our calibration can increase the separability of confidence of correct and incorrect samples.
    
    \item We diagnose commonly used calibration loss functions including the CE and MSE loss: they are often limited under test samples of certain characteristics. %We show that the commonly used MLE method sometimes does not agree with this goal.
\end{itemize}

\section{Related Work}
\label{sec:related_work}

\textbf{Loss functions used in post-hoc calibration.} Existing post-hoc calibrators typically use the Maximum Likelihood Estimation (MLE) for optimization~\citep{jung2023scaling,tao2023dual}. For example, \citet{mukhoti2020calibrating} use the CE loss, while \citet{kumar2019verified} use the MSE loss. Additionally, \citet{mukhoti2020calibrating} uses the focal loss, a variant to CE, which enhances learning on wrong predictions. \citet{guo2017calibration} directly optimizes the Expected Calibration Error (ECE). % Post-hoc calibration adjusts the outputs of a pre-trained neural network by modifying the confidence scores \citep{guo2017calibration, tomani2021post, tomani2023beyond, xiong2023proximity}. For example, vector scaling and matrix scaling \citep{guo2017calibration}, extended from Platt scaling, introduce a linear layer optimized by the NLL loss to transform the network logit vector, thereby calibrating its outputs. Differently, Spline \citep{gupta2020calibration} employs spline-fitting to create a recalibration function that directly maps the classifier outputs to calibrated probabilities. Dirichlet \citep{kull2019beyond} introduces a multi-class calibration method derived from Dirichlet distributions, while \citet{rahimi2020intra} propose a general post-hoc calibration function capable of preserving the top-$k$ predictions of any deep network through intra-order-preserving functions. % \blue{
%These popular post-hoc calibrators typically use the maximum likelihood estimation (MLE) 
%(\eg, minimizing cross-entropy loss or mean square error)
%for optimization, such as the cross-entropy loss and mean squared error.
This paper identifies inherent problems with MLE for calibration and derives a new loss function that better aligns with calibration goal. 
%This paper investigates an alternative loss function designed to effectively align with a mathematically derived calibration objective and how to make it work in practice. %This objective emphasizes accurate predictions with high confidence and incorrect predictions with low confidence. 
%\james{Good to discuss conformal prediction literature and how the approach here differs.}
%, while we explore an alternative loss function that well aligns a mathematically derived calibration objective: correct prediction towards high confidence and wrong prediction towards low confidence.
% }

\textbf{OOD calibration} deals with distribution shifts in test sets \citep{tomani2021post}.
%focuses on model calibration under 
A useful practice is modifying the calibration set to cover OOD scenarios \citep{tomani2021post, krishnan2020improving}, but these methods 
%For example, Tomani \emph{et al.} \citep{tomani2021post} generalize existing post-hoc calibration methods by transforming the calibration set.
%Krishnan \emph{et al.} \citep{krishnan2020improving} introduce a new loss to limit the number of accurate and uncertain samples and inaccurate and certain samples. These methods 
are usually designed for specific OOD scenarios which may lead to compromised IND calibration performance. 
Another approach adapts to test data distribution, whether OOD or IND, using domain adaptation  % TranCal 
\citep{wang2020transferable}, calibration sets from multiple domains \citep{gong2021confidence}, or improving calibration sets by estimating test set difficulty \citep{zou2023adaptive}. %utilizes domain adaption to improve temperature scaling, while requiring re-training of the adaption model when giving a new test set.
% Gong \emph{et al.} 
%\citet{gong2021confidence} use calibration sets from multiple domains.
%ACE \citep{zou2023adaptive} uses test batches to decide test set difficulty, which weighs  calibrators pretrained on calibration sets of extreme difficulty levels. %creates two calibration sets with extreme calibration difficulty, and then ensembles these two outputs of the learned two calibrators by the adaptive weighting scheme.% it is also effective to use test batches for  calibrator update \citep{zou2023adaptive}.
% \blue{
In comparison, our calibrator is trained on the \lei{IND} validation set \lei{only}, and does not need test batches for update, but still demonstrates improved and competitive OOD calibration performance. 
%In comparison, our propose method does not modify the calibration or assume access to test batches. By modifying the loss function and incorporate neighboring features of each sample, we show that a superior OOD calibrator can be obtained.
% }

%\textbf{Adaptive calibration.}

%A few works \citep{wang2020transferable,zou2023adaptive} develop calibrators that adapt to test data distribution and thus continuous improvement under all scenarios.

%Our method \red{xxx}.

\textbf{Predicting classification correctness} has not been widely studied. Among the few, %is a less studied problem compared with calibration. % Xia \emph{et al} 
\citet{xia2023window} investigate a three-way classification problem: classify a test sample into correct prediction, wrong prediction, or an out-of-distribution sample (its category is outside the training label space). We find the task of predicting classification correctness closely connected to model calibration. %This work integrates prediction correctness into the model calibration problem, where during post-hoc training we enforce the calibrator to output high confidence for correct prediction and vice versa. 
%We demonstrate that the resulting confidences not only exhibit low calibration errors but also aid users in making informed decisions.
% We show the resulting confidences are not only with low calibration errors but also help the users make decisions.

\section{Approach}
% \paragraph{Notations.}  Regular fonts are scalars; vectors are denoted by lowercase boldface letters, \eg, $\textbf{x}$; matrices by the uppercase boldface, \eg, {\textbf{M}}.

\subsection{Defining calibration error from its goal}
\label{approach:goal}

\textbf{Notations.} We study calibration under the multi-way classification problem. Regular fonts are scalars, \lei{\eg, $\tau$}; vectors are denoted by lowercase boldface letters, \eg, $\mathbf{x}$; matrices by the uppercase boldface, \eg, {\textbf{X}} for an image.\footnote{For simplicity, we omit three color channels.} \lei{$\idx{C}$ denotes an index set of integers $\{1,\cdots,C\}$, operator `$;$' and $\oplus$ concatenate vectors, \eg, $\oplus_{i\in\idx{M}}\vv_i\!=\![\vv_1; \cdots; \vv_M]$.} 
A classifier $f$ takes a \lei{$d$}-dimensional input $\mathbf{x} \in \mathbb{R}^\lei{d}$ and its corresponding label \lei{$y \!\in \!\idx{C}$} with $C$ classes which are sampled from the joint distribution $p(\mathbf{x},y)\!=\!p(y|\mathbf{x})p(\mathbf{x})$. \lei{We use $\equiv$ to denote the equivalence.}
The output of $f$ is denoted as $f(\mX)\!=\!(\hat{y}, \hat{c})$, where $\hat{y}$ and $\hat{c}$ denote the predicted class and \lei{maximum} confidence \lei{score}, respectively. % \james{}
%We consider the multi-class classification problem. The $D$-dimensional input variable $\mathbf{x} \in \mathbb{R}^D$ and its corresponding label $y \in \{1,2,\dots, C\}$ with $C$ classes are sampled from the joint distribution $\lambda(\mathbf{x},y)=\lambda(y|\mathbf{x})\lambda(\mathbf{x})$. The output of classifier $f$ is denoted as $f(\mathbf{x})=(\hat{y}, \hat{c})$, where $\hat{y}$ and $\hat{c}$ refer to the predicted class and the classifier's confidence, respectively.

\textbf{Calibration goal.} According to \citet{guo2017calibration}, the goal of model calibration is to ``align confidence with the accuracy of samples.'' Based on this, existing literature define perfect calibration as:
\begin{equation}
    \mathbb{P}(\hat{y}=y|\hat{c}=c)=c, \forall c \in [0,1].
    \label{eq:goal}
\end{equation}
% 
%In other words,  

\textbf{\lei{Our calibration error formulation.}} We interpret Eq. \eqref{eq:goal} as: for any predicted confidence $\hat{c}$, \lei{the} expected classification accuracy $\mathbb{E}^\text{acc}_{\hat{c}}$ of model $f$ on the conditional distribution $p(\mathbf{x} |\hat{c})$ should equal $\hat{c}$. Based on this interpretation, we write the calibration error of classifier $f$ as a function of $\hat{c}$:
% \begin{equation}
%     l_f(\hat{c}) = D (\hat{c} \| \mathbb{E}^\text{acc}_{\hat{c}}) = {\| \hat{c}-\mathbb{E}^\text{acc}_{\hat{c}} \|}=\| \hat{c}- \int \mathbb{I}\{ y_\mathbf{x}=\hat{y}_\mathbf{x}\}dp(\mathbf{x}|\hat{c}) \|,
%     \label{eq:l_f}
% \end{equation}
\begin{align}
    & l_f(\hat{c}) = D (\hat{c} \| \mathbb{E}^\text{acc}_{\hat{c}}) = {\| \hat{c}-\mathbb{E}^\text{acc}_{\hat{c}} \|} \nonumber \\
    & \quad \quad =\| \hat{c}- \int \mathbb{I}\{ y_\mathbf{x}=\hat{y}_\mathbf{x}\}dp(\mathbf{x}|\hat{c}) \|,
    \label{eq:l_f}
\end{align}
where $D$ denotes discrepancy measurement and $\| \cdot \|$ denotes a norm. We use $\ell_2$- distance in our implementation. The indicator function $\mathbb{I}$$\{ \cdot \}$ returns $1$ if the given condition (\lei{the prediction matches the ground truth label accurately}) is true; otherwise, it returns 0. % It reflects the posterior probability of the

Denoting the distribution of predicted confidence as $p(\hat{c})$ and its probability density function as $dp({\hat{c}})$, the expectation of calibration error\footnote{This differs from the expected calibration error (ECE) metric. \lei{The ECE metric uses discretized histogram bins, whereas our calibration goal employs the continuous form.}} of $f$ on $p(\hat{c})$ can be expressed as:
\begin{equation}
    \mathbb{E}_f=\int l_f(\hat{c}) dp(\hat{c}),
    \label{eq:E_f}
\end{equation}
where $l_f(\hat{c})$ is defined in Eq. \eqref{eq:l_f}. A model \lei{$f$} is considered \lei{to be} perfectly calibrated if $E_f=0$. Model calibration is to optimize a calibrator which reduces $\mathbb{E}_f$ as much as possible.

\subsection{Correctness-aware loss}
\label{approach:ca-loss}
%\footnote{Again it is a different concept from the ECE metric.}}
% \lei{\textbf{Our correctness-aware loss.}} 
In practice, \lei{the} distribution $p(\hat{c})$ is unknown, \lei{hence}, Eq. \eqref{eq:E_f} cannot be directly computed. To solve this, we approximate the calibration error $\mathbb{E}_f$ by replacing $p(\hat{c})$ with an empirical distribution, formed by assembling Dirac delta functions \citep{dirac1981principles} centered at each predicted sample confidence $\hat{c}$ computed from a given dataset $\mathcal{D}=\{(\mathbf{x}_i, y_i)\}^n_{i=1}$, where $n$ is the number of samples:
\begin{equation}
    dp{(\hat{c})}=\frac{1}{n}\sum_{i=1}^n \delta_{\hat{c}_i}(\hat{c}).
    \label{eq:d_lambda_p}
\end{equation}
Substituting Eq. (\ref{eq:l_f}) and Eq. \eqref{eq:d_lambda_p} into Eq. \eqref{eq:E_f}, the empirical calibration loss can be written as:
% \begin{align}
%     & \!\!\!\mathbb{E}_f^{emp}=\frac{1}{n}\sum_{i=1}^n l_f(\hat{c}_i) \nonumber \\ 
%     & \quad \quad \!\!= \frac{1}{n}\sum_{i=1}^n \| \hat{c_i}- \int \mathbb{I}\{ y_\mathbf{x}=\hat{y}_\mathbf{x}\}d\lambda(\mathbf{x}|\hat{c_i}) \|.
%     %& 
%     \label{eq:E_f_emp}
% \end{align}
\begin{equation}
\begin{split}
    \mathbb{E}_f^\text{emp} = \frac{1}{n}\sum_{i=1}^n \| \hat{c_i}- \int \mathbb{I}\{ y_{\mathbf{x}_\lei{i}}=\hat{y}_{\mathbf{x}_\lei{i}}\}dp(\mathbf{x}_\lei{i}|\hat{c_i}) \|.
\end{split}
    \label{eq:E_f_emp}
\end{equation}
However, Eq. \eqref{eq:E_f_emp} is still hard to compute because the probability density function $dp(\mathbf{x}_\lei{i}|\hat{c}_i)$ is not accessible in practice. As such, we further discretize Eq. \eqref{eq:E_f_emp}, where we assume a finite number of samples $\{ \mathbf{x}_{ij} \}^m_{j=1}$ in \lei{each} distribution $p(\mathbf{x}_\lei{i}|\hat{c}_i)$. Consequently, \lei{$\mathbb{E}_f^{\text{emp}}$} is approximated as:
\begin{equation}
\begin{split}
    \mathbb{E}_f^\text{emp} \approx \frac{1}{n}\sum_{i=1}^n  \| \hat{c_i}-  \frac{1}{m}\sum_{j=1}^m \mathbb{I}\{ y_{\mathbf{x}_{ij}}=\hat{y}_{\mathbf{x}_{ij}}\} \|.
\end{split}
    \label{eq:E_f_emp_emp}
\end{equation}
\lei{Note that} in practice, $m=1$,  because there is only one test sample for each $p(\mathbf{x}_\lei{i}|\hat{c}_\lei{i})$, \eg,  a dataset only has one test sample with confidence 0.52893.\footnote{We have a mild assumption that a test set has few, if any, duplicated images, where Eq. \eqref{eq:E_f_emp_emp} will also approximately hold.} Therefore, we define the correctness aware (CA) loss as the empirical calibration loss of classifier $f$ on a test set with $n$ samples:
\begin{equation}
   \mathcal{L}_\text{CA}=\frac{1}{n}\sum_{i=1}^n \| \hat{c}_i - \mathbb{I} \{ y_{\mathbf{x}_{i}}=\hat{y}_{\mathbf{x}_{i}} \} \|.
    \label{eq:E_f_emp_final}
\end{equation}
% \lei{\textbf{Intuition.}} % The intuition behind Eq. \eqref{eq:E_f_emp_final} is straightforward.

% \lei{As shown in Eq. \eqref{eq:E_f_emp_final}: }
% for a correct prediction, where $\mathbb{I} \{ y_{\mathbf{x}_{i}}=\hat{y}_{\mathbf{x}_{i}} \}=1$, the predicted confidence is better to be close to 1; \lei{and for} a wrong prediction, where $\mathbb{I} \{ y_{\mathbf{x}_{i}}=\hat{y}_{\mathbf{x}_{i}} \}=0$, the predicted confidence is encouraged to be close to 0.
% As such, Eq. \eqref{eq:E_f_emp_final} is named the Correctness-Aware (CA) loss, used in our calibrator \lei{(see Fig.~\ref{fig:3NT4})} in replace of existing loss functions such as the CE and MSE loss.  
\begin{figure*}[t]%
\centering
\includegraphics[width=\textwidth]{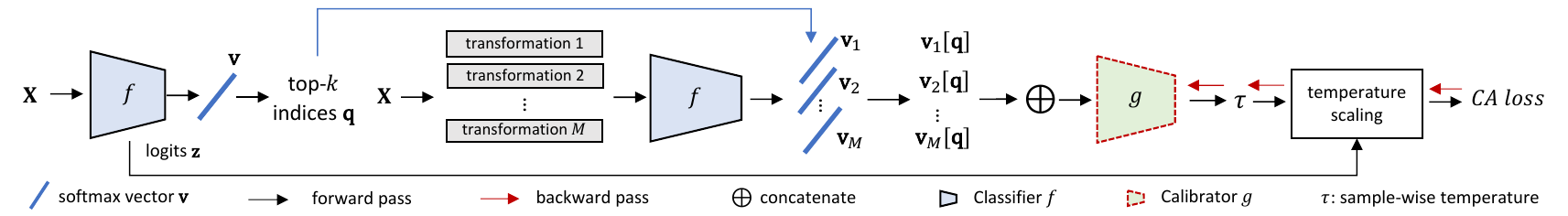}
\vspace{-5mm}
\caption{\textbf{Calibration pipeline.} For a given image sample $\mX$, we first obtain its logit vector $\mathbf{z}$ and softmax vector $\vv$. We then apply $M$ different transformations (\eg, rotation, greyscale, colorjitter, \etc) on $\mX$ to get its transformed versions as well as their related softmax vectors as $\vv_i$ ($i \in \idx{M}$). 
Indices $\vq\in\mbr{k}$ of the top-$k$ largest probabilities \lei{(softmax scores)} in $\vv$ are used to acquire top-$k$ scores from \lei{$\vv_i$} to form the concatenated input $\oplus_{i\in\idx{M}}\vv_i[\mathbf{q}]$ to the calibrator. \lei{The calibrator outputs} a temperature $\tau$, then being used to update the logit vector $\vz$ to produce the calibrated softmax vector. 
\lei{We use our proposed Correctness-Aware (CA) loss (Sec.~\ref{approach:ca-loss}).}}  
\label{fig:3NT4}
\vspace{-3mm}
\end{figure*}

\lei{\textbf{Lower and upper bounds of CA loss.}
Let us derive the CA loss range of Eq.~\eqref{eq:E_f_emp_final} for a given test sample $\mathbf{x}_{i}$:
(i) If the sample is correctly classified, the indicator function $\mathbb{I}^{+} \{ y_{\mathbf{x}_{i}}=\hat{y}_{\mathbf{x}_{i}}\}=1$. The maximum confidence range for a correctly classified sample is $\hat{c}_i^{+}\in(\frac{1}{C}, 1]$, and the loss range would be $\| \hat{c}_i^{+} - \mathbb{I}^{+} \{ y_{\mathbf{x}_{i}}=\hat{y}_{\mathbf{x}_{i}} \} \|\in[0, 1-\frac{1}{C})$.
(ii) If the sample is wrongly classified, $\mathbb{I}^{-} \{ y_{\mathbf{x}_{i}}=\hat{y}_{\mathbf{x}_{i}} \}=0$ and $\hat{c}_i^{-}\in(\frac{1}{C}, 1]$, the loss range would be $(\frac{1}{C}, 1]$.
%Therefore, the CA loss range for a given test sample is $[0, 1]$ whether it is classified correctly or wrongly.
}
Let $\rho \in [0,1]$ be the ratio of correctly classified samples in $\mathcal{D}$. 
\lei{For the correctly classified samples, we have:
\begin{equation}
    0 \leq \sum_{i=1}^{\rho \times n} \| \hat{c}_i^{+} - \mathbb{I}^{+} \{ y_{\mathbf{x}_{i}}=\hat{y}_{\mathbf{x}_{i}} \} \| < \rho \times n \times \left(1\!-\!\frac{1}{C}\right), \label{eq:posbound}\\
\end{equation}
and for the wrongly classified samples, we have:
\begin{align}
    & (1-\rho) \!\times\!n\!\times\!\frac{1}{C}\!<\!\!\!\!\!\sum_{i=1}^{(1-\rho) \times n} \!\!\!\!\!\| \hat{c}_i^{-} \!-\!\mathbb{I}^{-} \{ y_{\mathbf{x}_{i}}=\hat{y}_{\mathbf{x}_{i}} \} \| \! \leq\!(1\!-\!\rho) \!\times \! n \label{eq:negbound}.
\end{align}
Combining Eq.~\eqref{eq:posbound} and~\eqref{eq:negbound} with Equation~\eqref{eq:E_f_emp_final}, we can derive the lower and upper bounds of CA loss:
\begin{align}
    % & \quad\quad \!\mathbb{E}_f^\text{emp}=\frac{1}{n}\sum_{i=1}^n \| \hat{c}_i - \mathbb{I} \{ y_{\mathbf{x}_{i}}=\hat{y}_{\mathbf{x}_{i}} \} \|, \nonumber \\
    & n \! \times \! \mathcal{L}_\text{CA}=  \sum_{j=1}^{\rho \times n} \| \hat{c}_j^{+} \!- \!\mathbb{I}^{+} \{ y_{\mathbf{x}_{j}}\!=\!\hat{y}_{\mathbf{x}_{j}} \} \| \nonumber \\ 
    & \quad \quad \quad \quad + \!\!\sum_{k=1}^{(1-\rho) \times n} \| \hat{c}_k^{-} - \mathbb{I}^{-} \{ y_{\mathbf{x}_{k}}\!=\!\hat{y}_{\mathbf{x}_{k}} \} \|,  \label{eq:sum_of_pos_neg}\\
    & %n \times (1-\rho) \times \frac{1}{C} =
    n \!\times \!\left(\!\frac{1-\rho}{C}\!\right)\leq n \!\times \!\mathcal{L}_\text{CA} \leq n\times\left(\frac{1-\rho}{C} + \frac{C-1}{C}\right), \nonumber \\
    & \quad\quad\quad\frac{1-\rho}{C} \leq  \mathcal{L}_\text{CA} \leq \frac{1-\rho}{C} + \frac{C-1}{C}, \label{eq:bounds}
\end{align}
where $C$ is the number of classes. 
% Therefore, the CA loss has a lower bound of $|\mathcal{D}| \times \left(\frac{1-\rho}{C}\right)$ and an upper bound of $|\mathcal{D}|\times\left(\frac{1-\rho}{C} + \frac{C-1}{C}\right)$. We notice that both lower and upper bounds are highly related to $\frac{1-\rho}{C}$, the fraction of misclassified sample ratio per class in $\mathcal{D}$.
Hence, the CA loss has a lower bound of $\frac{1-\rho}{C}$ and an upper bound of $\frac{1-\rho}{C} + \frac{C-1}{C}$. We observe that both lower and upper bounds are closely tied to $\frac{1-\rho}{C}$, representing the fraction of misclassified samples per class in whole $\mathcal{D}$.
}

% \yuchi{Shall we make this claim a theorem or lemma?}
\lei{
% \textbf{CA loss reduces the overlap between correct and incorrect predictions.} Below we show the proof. 
\textbf{Theoretical insights of CA loss.}
First, we rewrite Eq.~\eqref{eq:E_f_emp_final} as a sum of two CA loss components, $\mathbb{I} \{ y_{\mathbf{x}_{i}}=\hat{y}_{\mathbf{x}_{i}}\}$ equals $1$ and $0$ for correctly and incorrectly classified samples, respectively, and we use $\ell_1$-distance for simplicity:
\begin{equation}
   \mathcal{L}_\text{CA}=\frac{1}{n} \Bigg(\sum_{j=1}^{\rho \times n} (1 - \hat{c}_j^{+})\! + \!\sum_{k=1}^{(1-\rho) \times n} \hat{c}_k^{-} \Bigg), 
\end{equation}
where $\hat{c}_j^{+}$ and $\hat{c}_k^{-}$ denote, respectively, the maximum confidence scores for correctly and incorrectly classified samples. Suppose $\rho \geq 50\%$, we have:
\begin{align}
    & \mathcal{L}_\text{CA}\! = \!\frac{1}{n} \Bigg(\sum_{j=1}^{\!\!(1\!-\!\rho) \!\times \!n} (1 \!- \!\hat{c}_j^{+}) + \!\!\!\!\sum_{j=(1-\rho)n+1}^{\rho \times n}\!\!\!\! (1 \!- \!\hat{c}_j^{+}) + \!\!\!\sum_{k=1}^{(1-\rho) \times n} \!\!\hat{c}_k^{-} \Bigg) \nonumber \\
    % & \qquad = \frac{1}{n} \Bigg(\sum_{j=1}^{(1-\rho) \times n} (1 - \hat{c}_j^{+}) + \sum_{k=1}^{(1-\rho) \times n} \hat{c}_k^{-} \Bigg)  \nonumber \\
    % & \qquad\qquad\qquad  +  \frac{1}{n}  \sum_{j=(1-\rho)n+1}^{\rho \times n} (1 - \hat{c}_j^{+}) \nonumber \\
    % & \qquad = \frac{1}{n} \Bigg((1-\rho)n + \sum_{k=1}^{(1-\rho) \times n} \hat{c}_k^{-}  - \sum_{j=1}^{(1-\rho) \times n} \hat{c}_j^{+} \Bigg)  \nonumber \\
    % & \qquad\qquad\qquad  +  \frac{1}{n}  \sum_{j=(1-\rho)n+1}^{\rho \times n} (1 - \hat{c}_j^{+}) \nonumber \\
    & = 1\!-\!\rho \!+ \!\frac{1}{n} \Bigg(\!\!\sum_{k=1}^{(1-\rho) \times n} \!\!\! \hat{c}_k^{-}  \!- \!\!\!\!\sum_{j=1}^{(1-\rho) \!\times \!n} \!\!\!\hat{c}_j^{+} \!\!\Bigg)  \!+\!  \frac{1}{n}  \!\!\!\!\sum_{j=(1-\rho)n+1}^{\rho \times n} \!\!\!\!(1 \!- \!\hat{c}_j^{+}). \label{eq:rewrite}
\end{align}
Now using $\mathbb{E}^\text{diff}\!\!=\!\!\frac{1}{(1-\rho) \times n} \Big(\!\sum_{k=1}^{(1-\rho) \times n} \hat{c}_k^{-}  \!-\! \sum_{j=1}^{(1-\rho) \times n} \hat{c}_j^{+}\!\Big)$ to denote the expectation of the difference in maximum confidence values between correct and incorrect predictions, and $\mathbb{E}^\text{+}=\frac{1}{(2\rho-1)n}\sum_{j=(1-\rho)n+1}^{\rho \times n} (1 - \hat{c}_j^{+})$ to denote the expectation of the maximum softmax scores for the rest $(2\rho-1)n$ correctly classified samples, Eq.~\eqref{eq:rewrite} can be written as:
\begin{align}
    & \mathcal{L}_\text{CA}\!\!=\!\!(1\!-\!\rho)\mathbb{E}^\text{diff} \!\!+ \!\!(2\rho\!-\!1)\mathbb{E}^\text{+} \!+ \!(1\!-\!\rho).
\end{align}
We notice that minimizing our CA loss is equivalent to minimizing either $\mathbb{E}^\text{diff}$ or $\mathbb{E}^\text{+}$ (omitting the constant $(1-\rho)$): (i) minimizing $\mathbb{E}^\text{diff}$ aims to maximize the expectation of the difference in maximum confidence scores between correct and incorrect predictions, and (ii) minimizing $\mathbb{E}^\text{+}$ aims to push the maximum confidence score of correctly classified samples to 1. These two objectives align well with the model calibration goal (Sec.~\ref{approach:goal}). Below we take a close look at $\mathbb{E}^\text{diff}$:
\begin{align}
    & \mathbb{E}^\text{diff}\!=\!\frac{1}{(1\!-\!\rho) \!\times \!n} \Bigg(\!\sum_{k=1}^{(1-\rho) \times n} \hat{c}_k^{-} \! - \!\sum_{j=1}^{(1-\rho) \times n} \hat{c}_j^{+}\Bigg)  \nonumber \\
    & \quad \quad \!\!\! =  \frac{1}{1-\rho}\Bigg(\mathcal{L}_\text{CA} - \frac{1}{n} \sum_{j=(1-\rho)n+1}^{\rho \times n} \!\!\!(1 - \hat{c}_j^{+}) + \rho -1\Bigg), \nonumber \\
    & \mathcal{L}_\text{CA} \rightarrow \frac{1-\rho}{C} \equiv \mathbb{E}^\text{diff} \rightarrow \frac{\Big(\frac{1}{C} - 1\Big)\rho}{1-\rho} < 0.
    \label{eq:overlap}% \\
    %& \!\!\!\!\!\!= n \!\times\! \Bigg(\mathbb{E}_f^\text{emp} - \frac{1}{n} \sum_{j=(1-\rho)n+1}^{\rho \times n} \!\!\!(1 - \hat{c}_j^{+}) + \rho -1\Bigg) \nonumber \\
    % & 
\end{align}
Eq.~\eqref{eq:overlap} demonstrates that minimizing our CA loss $\mathcal{L}_\text{CA}$ during training toward the lower bound $\frac{1-\rho}{C}$ is equivalent to pushing $\mathbb{E}^\text{diff}$ toward $\frac{\big(\frac{1}{C} - 1\big)\rho}{1-\rho} < 0$. This means pushing the average maximum softmax scores of wrongly classified samples away from those of correctly classified samples, thereby reducing the overlap of confidence values between correct and incorrect predictions.
}

% 

% Because post-hoc calibration is accuracy-preserving, we immediately derive a lower bound of $\mathbb{E}_f^\text{emp}$ from Eq. \eqref{eq:E_f_emp_final}:
% \begin{equation}
%     \inf_{f} \mathbb{E}_f = \sum_{i=1}^{\rho \times |\mathcal{D}|} \|\frac{1}{C}\| + \sum_{i=1}^{(1-\rho) \times |\mathcal{D}|} 0 ,
%     \label{eq:low_bound}
% \end{equation}
% where $C$ is the number of classes. Under this lower bound scenario, the confidence of correctly and wrongly classified samples is 1 and $\frac{1}{C}$, respectively. 

% a joint distribution of $\lambda(\mathbf{x}, \hat{c})=\lambda(\mathbf{x}|\hat{c})\lambda(\hat{c})$, the expected calibration error of $f$ on the this joint distribution can be obtained as the following derivation:
% \begin{equation}
% \begin{split}
%     \mathcal{E}_f = \int {\| \hat{c}-E^{acc}_{\hat{c}} \|} d\lambda(\hat{c}) \\
% = \int \int{\| \hat{c}- \mathbb{I}\{ y_\mathbf{x}=\hat{y}_\mathbf{x}\} \|} d\lambda(\mathbf{x}|\hat{c}) d\lambda(\hat{c}) \\
% =\int {\| \hat{c}- \mathbb{I}\{ y_\mathbf{x}=\hat{y}_\mathbf{x}\} \|} d\lambda(\mathbf{x}, \hat{pi})
% \end{split}
% \end{equation}
% Consequently, the expected calibration error of classifier $f$ on distribution $\lambda$ $\mathcal{E}_{cal}$ can be defined as the integral of the discrepancy between the variable $\hat{c}$ and the associated accuracy expectation $\mathbb{E}_{A}$:
% \begin{equation}
%     \mathcal{E}_{cal}=(\hat{y}=y|\hat{c}=p)=p, \forall p \in [0,1].
% \end{equation}

\subsection{Gaining correctness awareness}
\label{Sec:network}
From the intuition of the CA loss in Eq. \eqref{eq:E_f_emp_final}, its optimization requires the post-hoc calibrator to be aware of the correctness of each test sample. %This is a non-trivial task. %In our empirical 
%We consider minimizing the empirical calibration loss defined in Eq. \eqref{eq:E_f_emp_final} as our optimization objective for calibrator training.
%Intuitively, it encourages the calibrator to assign high confidence to correctly classified samples and low confidence to incorrectly classified ones. In other word, this optimization explicitly push the calibrator to be aware of the correctness of predictions. 
Empirically, we find the test sample itself offer limited help to distinguish correctness, which leads to undesirable calibration performance. % correctness only based on  itself resulting undesired calibration results in the testing stage (\eg, (improving) decreasing confidence for (in)correct samples).

% Our solution to gaining correctness awareness is motivated by \citep{deng2022strong}. They find consistency of model predictions of transformed (\eg, rotated, gray-scaled, \etc) images is highly correlated with model accuracy. While the insight from \citep{deng2022strong} is on the dataset level, our assumption is on the individual sample level: model behavior on transformed samples is useful to inform the correctness of model predictions. 

Our approach, inspired by \citep{deng2022strong}, leverages the discovery that consistency in model predictions for transformed images correlates strongly with accuracy. While their insight focuses on dataset-level consistency, our assumption extends to individual samples: the model's behavior on transformed samples informs prediction correctness.

%\textbf{Sample-wise temperature.} To solve this problem, we are inspired by the recent success \citep{deng2022strong, deng2021does} of using image transformations in unsupervised model evaluation. We hold that model behaviors on sample transformations may be informative cues to tell the correctness of model predictions and leading desired confidence calibration.

% \begin{algorithm}
% \caption{Get the calibrator input}
% \label{algorithm_1}
% \begin{algorithmic}[1]
% \Require $\mathbf{v}$ (Softmax vector of the original image), $\{\mathbf{v}_1, \mathbf{v}_2, \dots, \mathbf{v}_n\}$ (softmax vectors for $n$ transformed images), $n$ (number of transformed images), $k$ (number of largest values)
% \Ensure $(n \times k)$-dimensional vectors for calibrator input

% \Procedure{ExtractAndConcatenate}{$\mathbf{v}$, $\mathbf{v}_1, \mathbf{v}_2, \mathbf{v}_3, \dots$, $k$}
%     \State $I_k \gets \text{Indices of the } k \text{ largest values in } \mathbf{v}$
%     \If{$|I_k| < k$}
%         \State \textbf{raise} Error("Not enough values in $\mathbf{v}$ to extract $k$ indices.")
%     \EndIf
%     \State Initialize $V$ as an empty vector
%     \For{each transformed vector $\mathbf{v}_i$}
%         \State Initialize $v_{k}^{i}$ as an empty vector
%         \For{each index $j$ in $I_k$}
%             \State Append $\mathbf{v}_i[j]$ to $v_{k}^{i}$
%         \EndFor
%         \State $V \gets V \oplus v_{k}^{i}$ \Comment{$\oplus$ denotes vector concatenation}
%     \EndFor
%     \State \Return $V$
% \EndProcedure
% \end{algorithmic}
% \end{algorithm}

\lei{The pipeline of our calibrator is presented in Fig. \ref{fig:3NT4}}, we aim to calibrate a classification model \lei{$f$}. To do so, we compute \lei{the logit vector $\vz$, softmax vectors of an original image $\mX$ and its transformed versions $\vv$ and $\vv_i$ ($i \in \idx{M}$, assuming $M$ types of transforms), respectively.} We determine the indices $\mathbf{q}\in\mbr{k}$ of the $k$ largest \lei{softmax scores} of $\vv$, and use these indices $\vq$ to locate and select the corresponding values from $\vv_i$, forming new vector with $k$ dimensions. These $k$-dim vectors $\vv_i[\mathbf{q}]$ of the transformed images are \lei{concatenated as $\oplus_{i\in\idx{M}}\vv_i[\mathbf{q}]$, and} used as the calibrator input. %\yuchi{We can find the description of the calibrator input formulation in Algorithm \ref{algorithm_1}}.

Calibrator, \lei{$g$ parameterized by $\theta$}, which is trained by the proposed CA loss. Here, the calibrator consists of two fully connected layers with a ReLU activation function in between. Each hidden layer comprises 5 nodes. % To be specific, we use the concatenated softmax vectors of different transformations of a given image $[\oplus_{i\in\idx{M}}\vv_i]$ as the calibrator inputs. 
% \begin{equation}
% \arg\min_{\theta} \; \mathbb{E}_f^\text{emp} (\vv, \vv_1, \vv_2, \vv_3, \cdots),
% \label{eq:optim_goal}
% \end{equation}
% $\vv$ is the softmax vector of the original image, and $\vv_i$ and $i \in \idx{M}$ are those of its transformed versions. 
The calibrator is optimized on the calibration (\aka, validation) set, and the calibration output is temperature \lei{$\tau$} to be used to scale the model logit vector \lei{$\vz$ of original image. Below we show these steps in equations:
\begin{align}
    & \mathbf{q} = \underset{\mathbf{q}}{\text{argmax}} \vv \in\mbr{k}, \label{eq:topkk}\\
    & \tau = g_\theta(\oplus_{i\in\idx{M}}\vv_i[\mathbf{q}]), \label{eq:concate}\\
    & \hat{c} = \underset{c}{\text{max}}\sigma(\vz / \tau)^{(c)}, \label{eq:max}
\end{align}}
\lei{where $\sigma(\cdot)$ denotes the softmax function. Eq.~\eqref{eq:max} retrieves the maximum softmax prediction score $\hat{c}$.}
Based on the CA loss in Eq. \eqref{eq:E_f_emp_final}, the optimization goal now becomes:
\lei{\begin{equation}
\arg\min_{\theta}\mathcal{L}_\text{CA} (\hat{c}, y_\vx, \hat{y}_\vx), \label{eq:optim_goal}
\end{equation}}
% where $\theta$ is the parametric calibrator $g$ and the 
\lei{where $y_\vx$ and $\hat{y}_\vx$ denote respectively the ground truth and predicted labels.}% ,$f$ is the classification model we want to calibrate.

In practice, transformations can be grayscale, rotation, color jitter, adding Gaussian noise, random erasing, \etc. They are applied to the original image during training and inference. We do not assume access to test batches, which are used in some previous works~\citep{guo2017calibration, wang2023adaptive}. Nevertheless, if we assume such access, we can retrieve from test batches images that are similar to the original one and use softmax vectors \lei{of} the retrieved images as calibrator inputs. As to be shown in Sec.~\ref{sec:further-analysis}, we find that grayscale, rotation, and colorjitter are an effective combination, and that using three transformations give a good trade-off between calibration performance and computational cost. 

%Specifically, we employ three different transformations (rotation, grayscale, and colorjitter) on a image. The predictive probability vectors on this image and its transformed versions are notated as $\vv, \vv^r, \vv^g$ and $\vv^c$, respectively.
%To make the calibrator input is correctness informative, we select the probabilities in $\vv^r, \vv^g, \vv^c$ on the classes that has the top-$k$ probability in $\vv^r$. Intuitively, more similar the probabilities on these selective classes are with the probabilities on corresponding classes, more possible is image is classified correctly.
%Then, we concatenate those sleeted probability values into a single vector as the calibrator input feature.
%In practice, we make $k=4$ as the default configuration thus leading a input feature with 12 probability values from three different transformed images. The feature formulation procedure is illustrated in Fig. \ref{fig:3NT4}.

%where $\theta$ is our parametric calibrator and the $f$ is the classification model we need to calibrate. \james{Notation is not clear. Is $f$ the calibrator or the base classifier? Is this calibrator loss optimized on the training set or the validation set? What is the output of the calibrator? Seems like the calibrator can change the base model's predicted class; is this a problem?}  \yuchi{Thanks. $f$ here was the calibrator. Our method is post-hoc so that it does not change the base model's predicted class. I will revise the paper to make it more clear.}

During inference, using the $k$-dim vectors from the transformed images, we obtain an adjusted temperature from the calibrator. 
% This temperature is used to scale the logits of the original image, the softmax vector of which is then updated. 
Alg. \ref{algorithm_1} in Appendix~\ref{suppl:alg} summarises this calibration process.

%The network structure of $\theta$ is straightforward, adhering to the practices outlined in PTS \citep{tomani2022parameterized}. 
% It comprises two fully connected layers intertwined with a ReLU in between, with the hidden layer consisting of 5 nodes. 
%It consists of two fully connected layers interconnected with a ReLU activation function in between. The hidden layer comprises 5 nodes.
%
%

%Unlike the PTS method, which uses the top-$k$ logits of the test image as input, we construct more informative features for distinguishing correctness as described above, and use these as the input for the calibrator. The output of our calibrator is the sample-wise temperature for scaling the model logit of the raw input.

\subsection{Comparison between CA loss and MLE}

\begin{figure*}[t]%
\centering
\vspace{-3mm}
\includegraphics[trim=0 0 0 0cm, clip=true,width=\textwidth]
{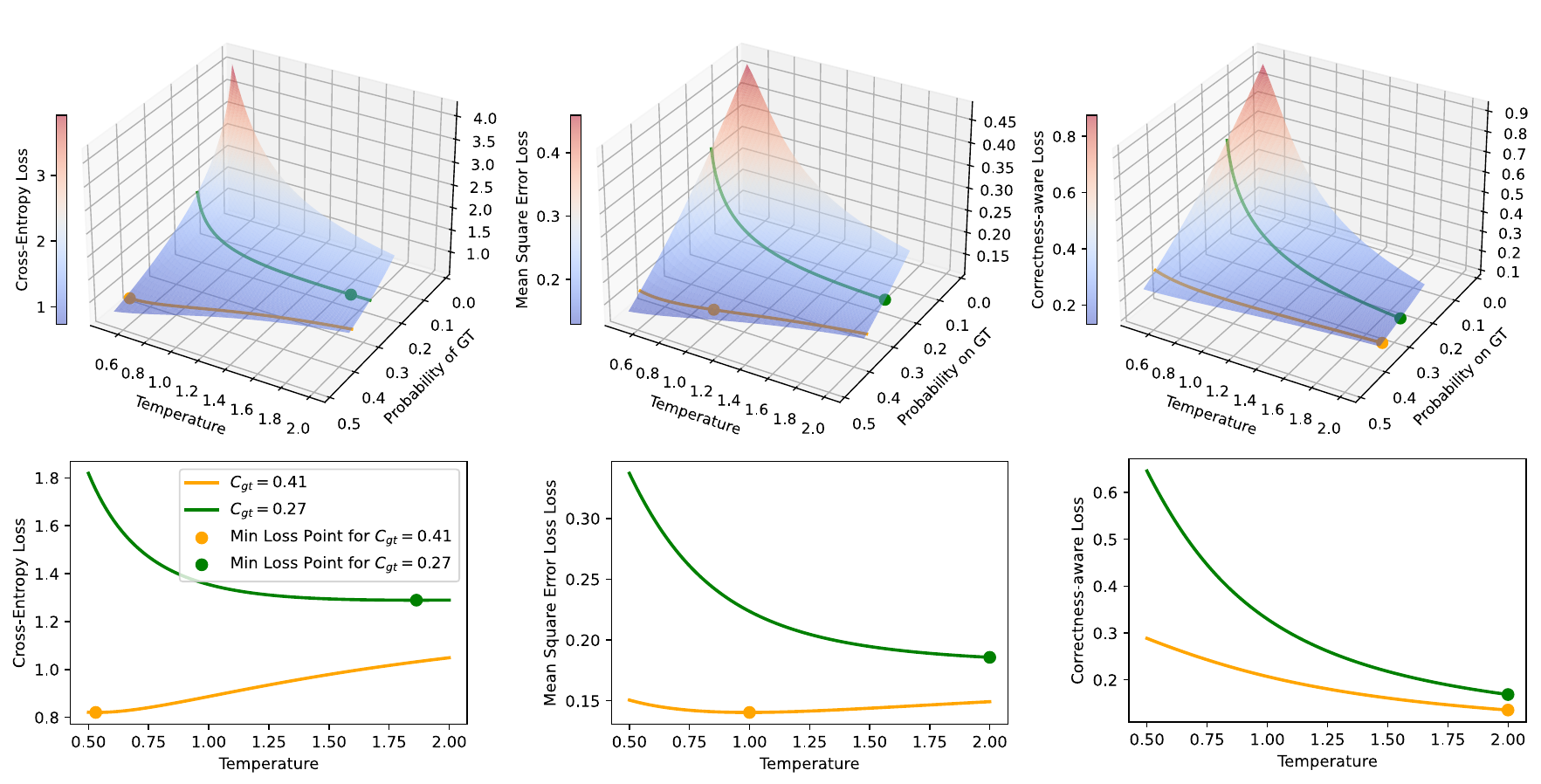}
\vspace{-8mm}
\caption{\textbf{Comparison of different loss functions \wrt  temperature and softmax probability of the ground truth (GT) class.} In a four-way classification task, we examine a wrongly predicted sample with logit vector $[a, 2.0, 0.1, 0.05]$, where $a\!<\!2$ is the value on the ground truth class. % and $a<2$. 
We use \lei{$c_\text{gt}$} to donate the softmax score of the GT class.
\textbf{Top:} The loss \lei{surface plots} for varying temperatures and $c_\text{gt}$, with red and blue arrows representing positive and negative temperature gradients, respectively. \textbf{Bottom:} Shows 2D loss curves for varying $c_\text{gt}$. The lines in the bottom charts correspond to the lines of the same color in the top charts. Compared with Maximum Likelihood Estimation (MLE) based functions (\eg, Cross-Entropy, Mean Squared Error), our Correctness-Aware loss minimization does not favor temperatures below 1 for incorrect predictions, while sometimes MLE does. }  
\label{fig:loss_surface}
\vspace{-3mm}
\end{figure*}

When a sample is correctly predicted, both CA loss and MLE (e.g., cross-entropy, mean squared error) behave similarly, pushing the ground-truth probability toward 1 and thereby improving calibration.
However, for wrong predictions, MLE can inadvertently raise confidence for narrowly wrong samples. As shown in Fig. \ref{fig:loss_surface}, for a narrowly wrong prediction (yellow curve), MSE and Cross-Entropy minimize their loss at a temperature greater than 1—contradicting the goal of lowering confidence for incorrect predictions. In contrast, CA loss steadily decreases confidence in such cases, resulting in better overall calibration. Please see Appendix \ref{sec:theory_comparison} for more detailed discussion.

% For narrowly wrong predictions (where the correct class’s probability is relatively close to the wrong class’s probability), MLE can unintentionally increase the model’s confidence instead of lowering it. This deviates from the calibration goal, as we do not want a model to be confident in an incorrect prediction. In contrast, CA loss consistently lowers the confidence for such samples, aligning with proper calibration.

\begin{table*}[t]
\centering
\setlength{\tabcolsep}{0.85mm}{
    
    \caption{\textbf{Calibrator comparison under the ImageNet setup.} Each reported number is averaged over 10 classifiers, %. Experiments are conducted on 10 classifier, as 
    described in Sec.~\ref{Sec:exp_setup}. We use five test sets: ImageNet-Val, ImageNet-A, ImageNet-R, ImageNet-S, and ObjectNet, %Each result represents an average across 10 classifiers. 
    and four metrics: ECE (bin=25), BS, KS and AUC (AUROC). Best results in each column are in bold. When comparing CA and CE, better results are in blue color. 
    }
    \vspace{-0.2cm}
    \label{tab:imagenet_setup}
\scriptsize
    \begin{tabular}{l cccc cccc cccc cccc cccc}
            \toprule
            \multirow{2}{*}{\textbf{Method}} & \multicolumn{4}{c}{ImageNet-Val}  & \multicolumn{4}{c}{ImageNet-A} & \multicolumn{4}{c}{ImageNet-R} & \multicolumn{4}{c}{ImageNet-S} & \multicolumn{4}{c}{ObjectNet} \\
            \cmidrule(lr){2-5} \cmidrule(lr){6-9} \cmidrule(lr){10-13} \cmidrule(lr){14-17} \cmidrule(lr){18-21} 
            & ECE$\downarrow$ & BS$\downarrow$ & KS$\downarrow$ & AUC$\uparrow$ & ECE$\downarrow$ & BS$\downarrow$  & KS$\downarrow$ & AUC$\uparrow$ & ECE$\downarrow$ & BS$\downarrow$ & KS$\downarrow$ & AUC$\uparrow$ & ECE$\downarrow$ & BS$\downarrow$ & KS$\downarrow$ & AUC$\uparrow$ & ECE$\downarrow$ & BS$\downarrow$ & KS$\downarrow$ & AUC$\uparrow$\\
            \midrule
            Uncal & 8.71 & 14.01 & 14.39 & 85.96 & 39.44 & 32.90 & 43.47 & 61.87 & 13.97 & 16.89 & 19.90 & 88.06 & 20.92 & 21.67 & 29.01 & 82.57 & 31.21 & 25.20 & 36.48 & 78.05 \\
            TS & 9.10 & 13.85 & 13.26 & 85.41 & 29.24 & 23.24 & 32.50 & 62.86 & 6.28 & 13.80 & 14.51 & 88.27 & 8.92 & 16.11 & 19.18 & 83.22 & 19.70 & 17.97 & 26.92 & 78.35 \\
            ETS & 3.22 & 12.61 & 12.43 & 85.90 & 32.40 & 26.21 & 36.98 & 62.68 & 8.29 & 14.21 & 17.09 & 88.22 & 14.79 & 17.47 & 23.72 & 83.17 & 25.13 & 20.61 & 31.11 & 78.28 \\
            MIR & 2.36 & 12.51 & 12.93 & 85.92 & 34.70 & 26.92 & 39.37 & 61.87 & 10.33 & 14.71 & 19.43 & 88.01 & 17.39 & 18.28 & 26.41 & 82.55 & 27.56 & 21.29 & 33.48 & 78.02 \\
            SPL & 2.38 & 12.50 & 12.75 & 85.94 & 33.61 & 26.71 & 38.28 & 61.87 & 9.41 & 14.58 & 18.33 & 88.02 & 16.48 & 18.09 & 25.45 & 82.56 & 26.44 & 21.09 & 32.27 & 78.37 \\
            Adaptive TS & 6.35 & 14.20 & 11.04 & 82.94 & 29.30 & 23.61 & 32.99 & 61.43 & 5.65 & 14.29 & 14.68 & 86.99 & 9.65 & 16.96 & 20.01 & 81.01 & 20.77 & 18.99 & 27.79 & 76.95 \\ 
            TCP & 8.30 & 17.38 & 17.45 & 72.15 & 28.07 & 19.62 & 32.05 & 47.57 & 8.81 & 22.59 & 21.16 & 62.77 & 9.79 & 21.69 & 25.56 & 56.95 & 21.79 & 18.58 & 31.25 & 72.57 \\ 
            ProCal & 2.89 & 12.60 & 13.33 & 86.08 & 38.82 & 30.99 & 42.35 & 61.75 & 11.84 & 16.29 & 19.13 & 86.22 & 19.85 & 19.63 & 28.10 & 82.30 & 25.22 & 22.61 & 31.58 & 75.15 \\
            \hline
            CE only (PTS) & 5.02 & 12.50 & \cellcolor{blue!10}11.40 & 86.69 & 41.23 & 32.02 & 45.60 & 60.85 & 19.14 & 18.70 & 26.44 & \cellcolor{blue!10}87.05 & 14.39 & 17.01 & 24.82 & 82.75 & 32.89 & 24.85 & 38.18 & 77.77 \\
            CA only & \cellcolor{blue!10}2.22 & \cellcolor{blue!10}12.25 & 12.63 & \cellcolor{blue!10}86.74 & \cellcolor{blue!10}32.14 & \cellcolor{blue!10}25.28 & \cellcolor{blue!10}36.47 & \cellcolor{blue!10}61.08 & \cellcolor{blue!10}11.62 & \cellcolor{blue!10}15.79 & \cellcolor{blue!10}20.44 & 86.86 & \cellcolor{blue!10}5.49 & \cellcolor{blue!10}15.29 & \cellcolor{blue!10}14.97 & \cellcolor{blue!10}82.50 & \cellcolor{blue!10}22.09 & \cellcolor{blue!10}18.59 & \cellcolor{blue!10}28.99 & \cellcolor{blue!10}77.89 \\  
            \hline
            CE+trans. & \cellcolor{blue!10}\textbf{3.42} & 12.87 & 11.77 & 85.36 & 28.06 & 22.01 & 32.38 & 63.47 & 6.22 & 13.28 & 15.30 & 88.64 & 11.52 & 15.94 & 21.08 & 83.84 & 20.89 & 18.28 & 27.69 & 78.50 \\
            %\hline
            CA+trans. (ours) & 4.63 & \cellcolor{blue!10}\textbf{11.85} & \cellcolor{blue!10}\textbf{11.55} & \cellcolor{blue!10}\textbf{87.44} & \cellcolor{blue!10}\textbf{20.65} & \cellcolor{blue!10}\textbf{16.79} & \cellcolor{blue!10}\textbf{22.50} & \cellcolor{blue!10}\textbf{63.74} & \cellcolor{blue!10}\textbf{4.91} & \cellcolor{blue!10}\textbf{12.21} & 
            \cellcolor{blue!10}\textbf{10.12} & \cellcolor{blue!10}\textbf{90.22} & \cellcolor{blue!10}\textbf{4.00} & \cellcolor{blue!10}\textbf{13.83} & \cellcolor{blue!10}\textbf{13.12} & \cellcolor{blue!10}\textbf{84.87} & \cellcolor{blue!10}\textbf{10.33} & \cellcolor{blue!10}\textbf{14.59} & \cellcolor{blue!10}\textbf{18.72} & \cellcolor{blue!10}\textbf{79.25}
 \\
            \midrule
            CA + CE + trans & 4.24 & 13.15 & 11.53 & 84.91 & 26.54 & 20.85 & 30.80 & 63.85 & 5.63 & 12.86 & 14.57 & 88.96 & 9.50 & 15.15 & 19.52 & 84.31 & 21.42 & 18.79 & 28.02 & 78.14 \\

            \bottomrule

    \end{tabular}
    }
    \vspace{-0.2cm}
\end{table*}

\begin{table*}[t]
\centering
\setlength{\tabcolsep}{1.75mm}{

    \caption{\textbf{Calibrator comparison under the CIFAR-10 setup.} Each number is averaged over 10 classifiers (see Sec.~\ref{Sec:exp_setup}). We use one IND test set (CIFAR10.1) and three OOD test sets (CIFAR-10.1 , CINIC, and CIFAR-10-C). Other notations are the same as Table \ref{tab:imagenet_setup}.} %, each result is averaged across 10 classifiers and we report four calibration error measurements: ECE (bin=25), BS, KS and AUC (AUROC). The best results are highlighted. 
    \label{table:cifar10_setup}
    \vspace{-0.2cm}
\scriptsize

    \begin{tabular}{l cccc cccc cccc cccc}
            \toprule
            \multirow{2}{*}{\textbf{Method}}  & \multicolumn{4}{c}{CIFAR-10.1} & \multicolumn{4}{c}{Gaussian Blur} & \multicolumn{4}{c}{Defocus Blur} & \multicolumn{4}{c}{CINIC} \\
            \cmidrule(lr){2-5} \cmidrule(lr){6-9} \cmidrule(lr){10-13} \cmidrule(lr){14-17}
            & ECE$\downarrow$ & BS$\downarrow$  & KS$\downarrow$ & AUC$\uparrow$ & ECE$\downarrow$ & BS$\downarrow$ & KS$\downarrow$ & AUC$\uparrow$ & ECE$\downarrow$ & BS$\downarrow$ & KS$\downarrow$ & AUC$\uparrow$ & ECE$\downarrow$ & BS$\downarrow$ & KS$\downarrow$ & AUC$\uparrow$\\
            \midrule
            Uncal & 10.22 & 12.59 & 13.80 & 85.08 & 45.72 & 43.48 & 50.01 & 66.7 & 34.79 & 34.79 & 39.73 & 71.49 & 24.25 & 25.29 & 28.42 & 78.76 \\
            TS & 4.84 & 11.06 & 11.78 & \textbf{85.15} & 35.25 & 34.54 & 42.81 & 66.78 & 24.99 & 28.17 & 33.93 & 71.40 & 16.04 & 20.77 & 24.17 & \textbf{79.15} \\
            ETS & 2.77 & \textbf{10.78} & 10.87 & 85.08 & 30.04 & 30.84 & 39.41 & 66.69 & 19.98 & 25.69 & 39.73 & 71.23 & 11.48 & 19.13 & 22.15 & 79.25 \\ 
            MIR & \textbf{2.29} & 10.83 & 11.01 & 84.98 & 30.39 & 30.83 & 39.73 & 66.67 & 20.31 & 25.63 & 31.51 & 71.41 & 12.58 & 19.59 & 22.86 & 78.68 \\
            SPL & 3.04 & 10.89 & \textbf{10.61} & 85.04 & 29.12 & 30.57 & 38.49 & 66.72 & 19.64 & 25.53 & 30.46 & 71.48 & 11.88 & 19.51 & 22.03 & 78.72 \\
            Adaptive TS  & 4.17 & 11.29 & 11.74 & 83.40 & 22.07 & 26.58 & 34.44 & 65.97 & 12.15 & 23.31 & 27.27 & 70.27 & 10.73 & 19.48 & 21.99 & 77.76    \\
            TCP & 11.40 & 13.83 & 12.78 & 76.09 & 12.38 & 24.64 & 30.48 &  54.83 & 6.71 & 24.66 & 24.35 & 58.25 & 15.35 & 23.62 & 17.47 & 73.16 \\
            ProCal & 2.96 & 11.27 & 11.60 & 82.85 & 37.26 & 36.83 & 44.28 & 64.50 & 25.68 & 28.99 & 34.36 & 69.39 & 18.04 & 22.19 & 25.14 & 76.22\\
            \hline
            CE Only (PTS) & 2.92 & 10.85 & 11.26 & \cellcolor{blue!10}85.12 & 31.01 & 31.24 & 40.07 & \cellcolor{blue!10}67.28 & 21.01 & 25.86 & 31.80 & 71.95 & 13.69 & 19.92 & 23.27 & \cellcolor{blue!10}78.93 \\
            CA Only  & \cellcolor{blue!10}2.42 & \cellcolor{blue!10}10.87 & \cellcolor{blue!10}11.06 & 84.82 & \cellcolor{blue!10}29.41 & \cellcolor{blue!10}30.19 & \cellcolor{blue!10}39.05 & 67.11 & \cellcolor{blue!10}19.45 & \cellcolor{blue!10}25.17 & \cellcolor{blue!10}30.99 & \cellcolor{blue!10}\textbf{72.04} & \cellcolor{blue!10}12.65 & \cellcolor{blue!10}19.68 & \cellcolor{blue!10}22.93 & 78.55 \\ 
            \hline
            CE+trans. & 2.87 & 10.90 & 11.22 & 84.84 & 18.49 & 24.51 & 32.22 & 67.18 & 9.09 & 22.04 & 25.79 & 71.52 & \cellcolor{blue!10}\textbf{7.05} & \cellcolor{blue!10}\textbf{18.31} & 20.65 & \cellcolor{blue!10}79.03  \\
            %\hline
            CA+trans. (ours) & \cellcolor{blue!10}2.76 & \cellcolor{blue!10}10.79 & \cellcolor{blue!10}10.72 & \cellcolor{blue!10}\textbf{85.15} & \cellcolor{blue!10}\textbf{11.45} & \cellcolor{blue!10}\textbf{22.18}  & \cellcolor{blue!10}\textbf{27.91} &  \cellcolor{blue!10}\textbf{67.33} & \cellcolor{blue!10}\textbf{4.65} & \cellcolor{blue!10}\textbf{21.30} & \cellcolor{blue!10}\textbf{22.42} & \cellcolor{blue!10}71.57 & 7.27 & 18.34 & \cellcolor{blue!10}\textbf{15.90} & 78.91\\

            \bottomrule
    \end{tabular}
    }
    \vspace{-0.20cm}
\end{table*}

\section{Experiments}
\label{sec:exp}

% \begin{figure*}[t]%
% \centering
% \includegraphics[width=1\textwidth]{assets/roc_curve.pdf}
% \vspace{-1cm}
% \caption{\textbf{Comparison of Receiver Operating Characteristic (ROC) Curves Across Different Calibration Methods}. Each figure's title specifies the classifier and the test set used. It is evident that our methods (green curves) yield a higher Area Under Curve (AUC) compared to other calibration methods, signifying an enhanced ability of our model to distinguish between correct and incorrect predictions based on calibrated confidence.
% }  
% \label{fig:roc_conf}
% \end{figure*}

\subsection{Models and datasets}
\label{Sec:exp_setup}
\textbf{ImageNet-1k setup.}
\underline{1. Models.} We use 10 models  trained or fine-tuned on the ImageNet-1k training set~\citep{deng2009imagenet}. We source these models from the model zoo Timm~\citep{rw2019timm}. 
% As suggested by \citet{deng2022strong}, these models exhibit a diverse range of architectures, 
% (\eg, Convolution Neural Networks \citep{simonyan2014very, he2016deep, liu2022convnet}, Vision Transformers \citep{bao2021beit, dosovitskiy2020image, liu2021swin}, and MLP architectures \citep{ding2021repmlp, tolstikhin2021mlp}), 
% training strategies, and pre-training settings.
\underline{2. Calibration sets.} We use ImageNet-Val \citep{deng2009imagenet} to train calibrator.
\underline{3. Test sets.} 
(1) \emph{ImageNet-A(dversarial)}~\citep{hendrycks2021nae} comprises natural adversarial examples that are unmodified and occur in the real world. 
(2) \emph{ImageNet-S(ketch)}~\citep{wang2019learning} contains images with a sketch-like style.
(3) \emph{ImageNet-R(endition)}~\citep{hendrycks2021many} comprises of 30,000 images that exhibit diverse styles.
% like cartoons, deviant-art, graffiti, embroidery, and more.
(4) \emph{ObjectNet} \citep{barbu2019objectnet} is a real-world test set for object recognition where illumination, backgrounds and imaging viewpoints are very challenging.
(5) \emph{ImageNet-Val}. We train the calibrator on half of the ImageNet validation set and test it on the remaining half.
% (6) \emph{ImageNet-V2} \citep{recht2019imagenet} is a reproduced ImageNet dataset, whose distribution is similar to the ImageNet dataset.

\textbf{CIFAR-10 setup.}
\underline{1. Models.} 
We use 10 different models trained on the training split of CIFAR-10 \citep{krizhevsky2009learning} in this setup. We follow the practice in \citep{deng2022strong} to access the model weights. 
\underline{2. Calibration set.} Calibrators are trained on the test set of CIFAR-10.
\underline{3. Test sets} 
(1) \textit{CINIC-10} \citep{darlow2018cinic} is a fusion of \lei{both} CIFAR-10 and ImageNet-C \citep{hendrycks2019benchmarking} image classification datasets. It contains the same 10 classes as CIFAR-10. (2) \textit{CIFAR-10-C(orruptions)} contains subsets from CIFAR-10 modified by perturbations such as blur, pixelation, and compression artifacts at various severities. 

\textbf{iWildCam setup.} 
\underline{1. Model.} 
We use 10 models trained on the iWildCam\citep{beery2020iwildcam} training set. They are \lei{downloaded from} the \lei{official dataset} website. 
\underline{2. Calibration set.}  We train \lei{the calibrator} on the iWildCam validation set.
\underline{3. Test set.} We use the iWildCam test set containing animal pictures captured in the wild. Further details of the three setups are provided in Appendix~\ref{appendix-exp}.

\subsection{Calibration methods and evaluation metrics}
\textbf{Methods.} We compare our method with 8 popular calibration methods. They include scaling-based methods such as temperature scaling (TS) \citep{guo2017calibration}, ensemble temperature scaling (ETS) \citep{zhang2020mix}, adaptive temperature scaling (Adaptive TS) \citep{joy2023sample}, and parameterized temperature scaling (PTS) \citep{tomani2022parameterized}. We also compare with binning method multi-isotonic regression (MIR) \citep{zhang2020mix}, True
Class Probability (TCP) \citep{corbiere2021confidence}, spline-based re-calibration method (Spline) \citep{gupta2020calibration}, and Proximity-Informed Calibration (ProCal) \citep{xiong2024proximity}. 
% We use their official source codes for \lei{experiments}. 

\textbf{Metrics.} %We use various metrics to evaluate calibration performance. 
Apart from the expected calibration error (ECE) \citep{guo2017calibration}, we report the Brier score (BS), adaptive calibration error (ACE) \citep{nixon2019measuring}, and Kolmogorov-Smirnov (KS) error \citep{gupta2020calibration}.  %for comprehensive understanding of calibration performance. 
In addition, we use area under the ROC curve (AUROC) to evaluate how well the calibrators separate correct predictions from wrong predictions, which might also be a good metric for calibration (Appendix~\ref{sec:discussion}). All the numbers in Table \ref{tab:imagenet_setup},~\ref{table:cifar10_setup}, and~\ref{table:iwild_setup} are averaged over results of calibrating 10 different classification models (Appendix \ref{appendix-exp}). % introduced in Sec. \ref{Sec:exp_setup}. 

\subsection{Main observations}

% We offer additional analysis, including hyperparameter evaluation and computational cost analysis, in Appendix~\ref{sec:further-analysis}. Below we present our main observations.

%\textbf{CAMC make confidence score more coherent with the prediction correctness.} 

\begin{table}[t]
\vspace{-0.1cm}
\centering
\setlength{\tabcolsep}{4mm}{
    \caption{\textbf{Calibrator comparison under iWildCam setup.} Each number is averaged over 10 classifiers. 
    We use the iWildCam test set. Other notations are the same as Table \ref{tab:imagenet_setup}.}% Similar to Table \ref{tab:imagenet_setup}, each result is averaged across 10 classifiers and we report four calibration error measurements: ECE (bin=25), BS, ACE, and KS. The best results are highlighted.
    %} 
    \label{table:iwild_setup}
    \vspace{-0.2cm}
\scriptsize
    \begin{tabular}{l ccccc}
            \toprule
            \textbf{Method} & ECE$\downarrow$ & BS$\downarrow$ & KS$\downarrow$ & AUC $\uparrow$ \\
            \midrule
            Uncal & 16.04 & 17.77 & 22.11 & 86.59 \\
            TS & 6.63 & 14.06 & 14.90 & 86.22 \\
            ETS & 6.83 & 14.05 & 13.84 & 86.17 \\
            MIR & \textbf{5.01} & 13.53 & 12.57 & 86.61 \\ 
            SPL & 5.98 & 13.70 & 12.44 & 86.63 \\ 
            ProCal & 7.09 & 14.68 & 16.39 & 84.70  \\
            Adaptive TS & 7.17 & 14.25 & 12.88 & 86.33 \\
            TCP & 13.30 & 13.57 & 13.73 & 90.51 \\
            \hline
            CE only (PTS) & 8.73 & 17.90 & 17.45 & 78.65\\ 
            CA only & \cellcolor{blue!10}8.07 & \cellcolor{blue!10}17.54 & \cellcolor{blue!10}16.95 & \cellcolor{blue!10}79.15 \\  
            \hline
            CE+trans. & \cellcolor{blue!10}6.78 & 12.88 & 12.07 & 88.70 \\
            CA+trans. (ours) & 7.21 & \cellcolor{blue!10}\textbf{11.81} & \cellcolor{blue!10}\textbf{10.24} & \cellcolor{blue!10}\textbf{90.52} \\
            %\hline
                                  
            \bottomrule

    \end{tabular}
    }
\vspace{-0.4cm}
\end{table}

\textbf{Comparison of calibration performance with the state of the art.} %achieves state-of-the-art calibration performance on various out-of-distribution test sets.} 
We summarize calibration results under the ImageNet, CIFAR-10, and iWildCam setups in Table \ref{tab:imagenet_setup}, \ref{table:cifar10_setup}, and \ref{table:iwild_setup}, respectively. We have two observations. \textbf{First}, on OOD test sets, our method is very competitive across various metrics. For example, when compared with the second-best method on ObjectNet, the ECE, BS, and KS metrics of our method are 10.56\%, 3.69\%, and 8.97\% lower, respectively. \textbf{Second}, on near IND or IND test sets such as ImageNet-Val and CIFAR-10.1 , our method is less advantageous but is still competitive. The reason for our method being more effective on OOD test sets is that there are more narrowly wrong predictions, mentioned in the last paragraph in Sec. \ref{sec:theory_comparison}. Besides, as explained in Sec. \ref{Sec:network}, the use of transformed images might not be an optimal way to inform classification correctness. %in future better correctness prediction methods would improve the calibration performance of the  CA loss.

\textbf{Comparing CA with MLE.} In Table \ref{tab:imagenet_setup}, \ref{table:cifar10_setup}, and \ref{table:iwild_setup}, we compare `CA only' with `CE only', and `CA+trans'\footnote{`CA only' uses the CA loss and the top-$k$ logits of the original image as inputs to train the calibrator. `CA+trans' uses the CA loss and logits from transformed images instead (see, Fig.~\ref{fig:3NT4}).} with `CE+trans'. 
First, `CA only' consistently outperforms `MSE only' in 18 out of 20 scenarios under the ImageNet setup, 13 out of 16 scenarios under the CIFAR-10 setup, and 4 out of 4 scenarios under the iWildCam setup.  %across test sets. 
Second, in most cases (\eg, 19 out of 20 scenarios under ImageNet setup), `CA+trans' is better than `CE+trans'. 
In addition, in Table 1, we observe that the combination of CE and CA does not yield better results compared to using CA alone in the OOD test set. The superiority of the CA loss is more evident on OOD datasets as discussed in Appendix~\ref{sec:discussion}.

\begin{figure*}[t]%
\centering
\includegraphics[width=1\textwidth]{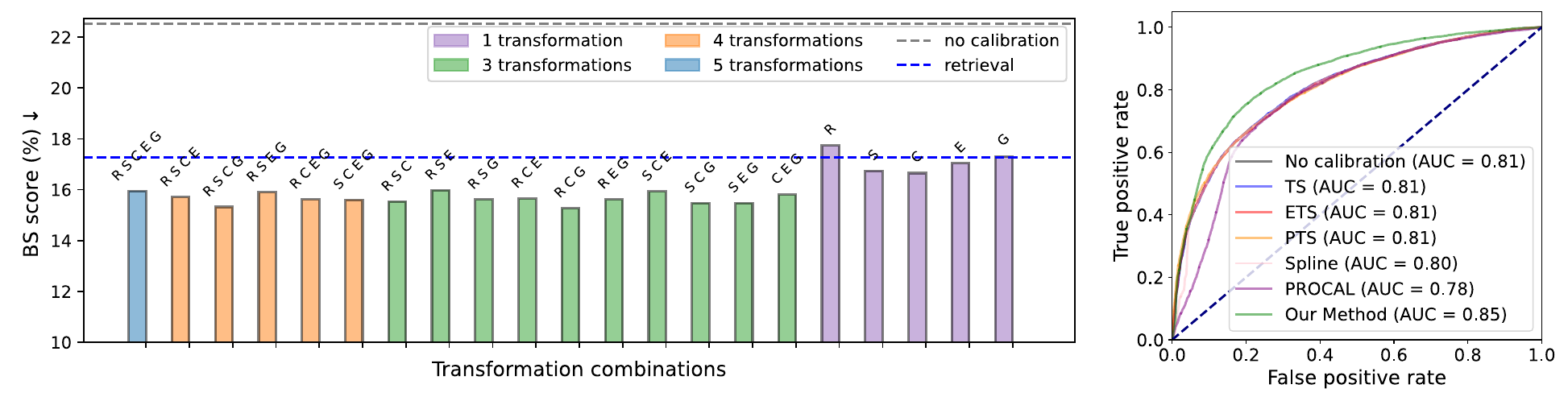}
\vspace{-6mm}
\caption{\textbf{(Left:)} Comparing various combinations of image transformations, including rotation (R), grayscale (S), colorjitter (C), random erasing (E) and Gaussian noise (G). Different colors means different numbers of transformations. Dashed lines denote performance of no calibration and retrieval-based augmentation that accesses test batches. \textbf{(Right:)} Visualization of ROC curves of various calibrators. Existing methods typically do not improve AUC, while our method effectively does. All results in this figure are reported for ObjectNet using the model `beit\_base\_patch16\_384', as introduced in Appendix \ref{appendix-exp}.
}  
\label{fig:roc_conf-trans}
\end{figure*}

\textbf{Potential of CA in allowing confidence to better separate correct and wrong predictions.} In Tables \ref{tab:imagenet_setup},  \ref{table:cifar10_setup}, and \ref{table:iwild_setup}, we compare separability and have two observations. First, existing methods typically do not have improvement in AUROC. This is not surprising, because their working mechanisms are not relevant to the separation of correct and wrong predictions. Second, our method improves AUROC under the ImageNet and iWildCam setups and in on par with existing methods on CIFAR-10. In fact, we find predictions of transformed images offer much less diversity under CIFAR-10 classifiers, losing their efficacy in telling prediction correctness. This could be addressed with a better method than transformed images, and we leave it for future work.  % using a stronger   has much fewer classes and that its classifer  % \blue{xxx}. 
These results, especially those under the challenging ImageNet and iWildCam setups, suggest our method has the potential to better distinguish between correct and wrong predictions by confidence scores, which could lead to improved decision making. A closer look at the ROC curves is provided in Fig.~\ref{fig:roc_conf-trans} and Appendix~\ref{suppl:vis}.

\begin{figure}[t]
\vspace{-0.1cm}
\centering
\includegraphics[width=0.49\textwidth]{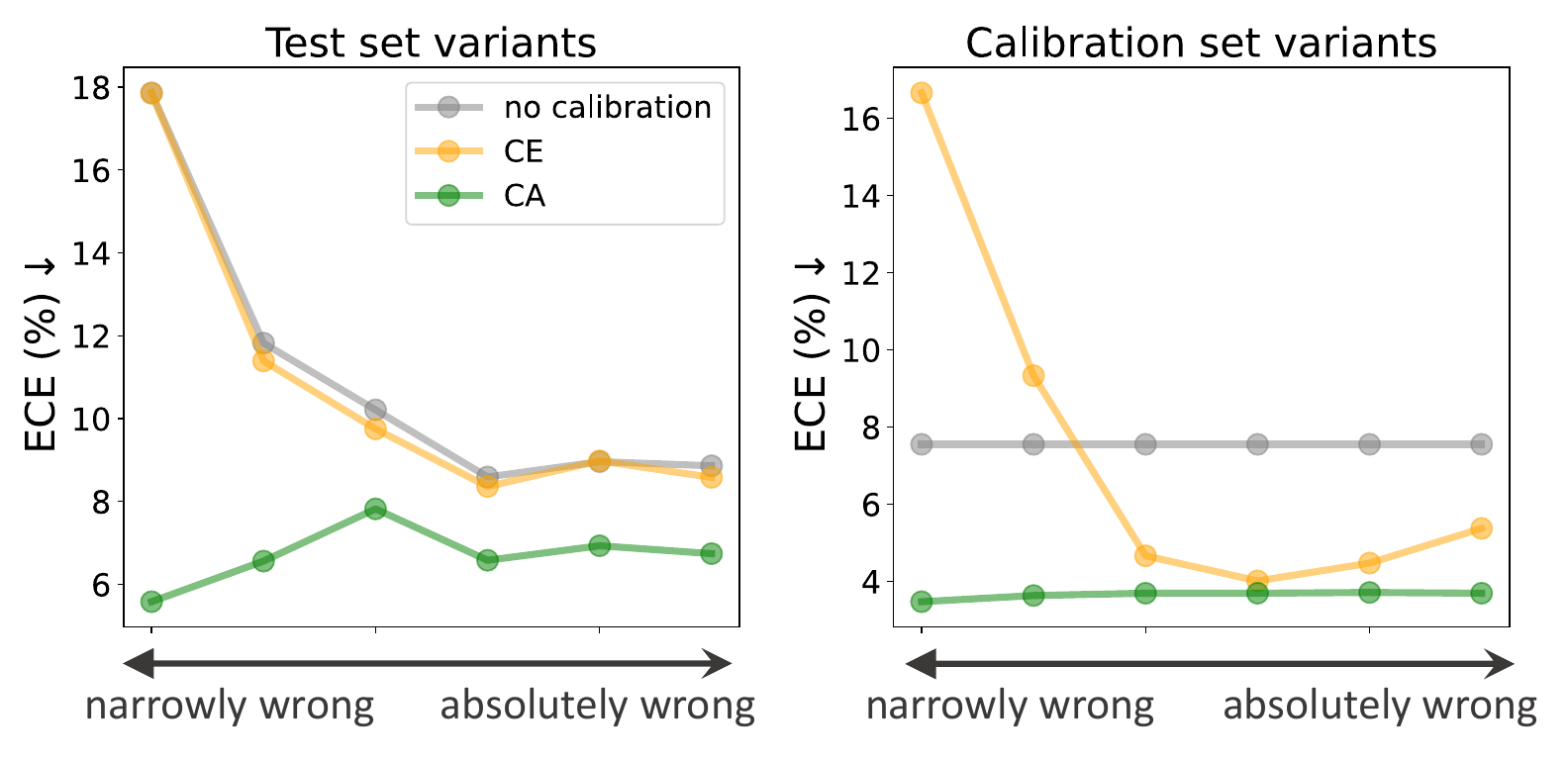}
\vspace{-6mm}
\caption{\textbf{Impact of narrowly wrong and absolutely wrong predictions on calibrator performance.} (\textbf{Left:}) we craft test sets containing 500 wrongly predicted samples with various degrees of being wrong. For example, the leftmost test set contains narrowly wrong samples, while the rightmost one contains absolutely wrong sample. Calibrator is trained on ImageNet-Val.   %$C_{gt}$ for wrong predictions in the training and testing.} We descending 
(\textbf{Right:}) we craft training sets containing 1,000 wrong predictions and 1,000 correct predictions. The wrongly predicted samples also have different degrees of being wrong. We use ImageNet-A as test set. 
For both subfigures, we use \textit{`beit\_base'} as the classifier and compare CA with CE and no calibration. Our method is more superior when training/test sets contain more narrowly wrong predictions. 
% 
%sort all wrong predictions of the 
%\textit{`beit\_base'} model according to their their confidence $C_{gt}$ on ground truth classes. 
%In testing (\textbf{left}), we combine 500 correct predictions and 500 wrong ones with controlled rank of $C_{gt}$ and report the ECE (\%) of the calibrator for \textit{`beit\_base'} trained from Table \ref{tab:imagenet_setup} in those test sets. In training \textbf{(right)}, we use 1000 correct ones and 1000 incorrect ones with different ranks to construct various calibration set. The ECE (\%) of calibrator variants are reported on ImageNet-A.
% r
}  
\label{fig:meta_set}
\vspace{-0.3cm}
\end{figure}

%\textbf{Decrease on calibrations errors for CIFAR-10 models in various ODD datasets.}

\textbf{Effectiveness of using transformed images as calibrator input.} We compare 
`CA+trans' and `CA only' in Table \ref{tab:imagenet_setup}, \ref{table:cifar10_setup}, and \ref{table:iwild_setup}. It is very clearly that `CA+trans' gives consistently better calibration performance than `CA only'. It indicates the necessity of using transformed images.

\section{Further analysis}
\label{sec:further-analysis}
%There are several potential factors that can impact the calibration performance of our method.

\textbf{Impact of narrowly wrong predictions in training and testing.} 
We construct various calibration (training) sets and test sets with samples of controlled degrees of being wrongly predicted. From Fig. \ref{fig:meta_set} (left), if a test set is dominantly filled with narrowly wrong predictions, our method will have a huge improvement over CE and no calibration: in fact, CE has the same performance as no calibration in this scenario. As more absolutely wrong samples are included, the gap between smaller, but our method is still superior. This is because the calibration set also has various degrees of wrong predictions, so CE is not as well trained as CA and actually has similar performance as no calibration. 

On the other hand, from Fig. \ref{fig:meta_set} (right), when training set contains lots of narrowly wrong predictions, CE is very poor and even worse then no calibration. When more absolutely wrong samples are included, CE becomes gradually better and even close to our method. These results empirically verifies our discussion in Sec. \ref{sec:theory_comparison}.

%values $C_{gt}$ (Fig. \ref{fig:loss_surface}) on ground truth classes in wrong predictions. According to the results in Fig. \ref{fig:meta_set}, we have two main observations. First, narrowly wrong predictions can have a significant negative impact on both calibrator training and testing when the calibrator is trained with cross-entropy (CE) loss. Second, CE can achieve a result close to CA if the calibrator is trained without narrowly wrong predictions, even though the test set contains such predictions (Fig. \ref{fig:meta_set} (right)). The negative impact of narrowly wrong predictions raised in training can be hard to eliminate in testing, even if we remove narrowly wrong predictions from the test set (Fig. \ref{fig:meta_set} (left)).

% In the test stage, the calibrator trained with our method (CA) has much lower ECE (3.47 \%) than that trained with CE (16.66\%) in the test set with narrowly wrong predictions.

\textbf{Comparing different image transformations}. We try different combinations of image transformations (including retrieval-based augmentation in test batches) as calibrator input. Results are summarized in Fig.~\ref{fig:roc_conf-trans}. We observe that using rotation, gray-scaling, and color-jittering generally give good calibration results. Retrieval-based augmentation is also competitive, but it requires access to test batches which might not be practical. Moreover, we find that using only one transformation is not ideal. While using more transformations is effective, three is a good number to balance between calibration performance and computational cost.

%\textbf{Impact of number of transformations.} We also try different numbers of transformation in Fig.~\ref{fig:roc_conf-trans} using more augmentation methods. Generally, the more transformations we use, the better calibration results we obtain. We choose three  considering the trade-off between calibration performance and computational cost.

%\textbf{Features from each type of image transformation are beneficial.} We study the impact of different image transformations used for extracting features to constitute the calibrator input. Specifically, we 

\section{Conclusion}
This paper starts from the general goal of calibration, mathematically interprets it, and derives a concrete loss function for calibration. Name as correctness-aware (CA) loss, in training it requires correct (wrong) predictions to have high (low) confidence, where such correctness is informed by transformed versions of original images. During inference, our calibrator also takes transformed images as input and tends to give high (low) confidence to likely correctly (wrongly) predicted images. We show our method is very competitive compared with the state of the art and potentially benefits decision making with plausible results on better separability of correct and wrong predictions. Moreover, we reveal the limitations of the CE and MSE losses for certain type of samples in the calibration set. Rich insights are given \textit{w.r.t} how our method deals with such samples. % in training and inference. 
In future we will study more effective correctness prediction methods to improve our system and how our method can be used for training large vision language models.

% \section*{Acknowledgements}

% \textbf{Do not} include acknowledgements in the initial version of
% the paper submitted for blind review.

% If a paper is accepted, the final camera-ready version can (and
% usually should) include acknowledgements.  Such acknowledgements
% should be placed at the end of the section, in an unnumbered section
% that does not count towards the paper page limit. Typically, this will 
% include thanks to reviewers who gave useful comments, to colleagues 
% who contributed to the ideas, and to funding agencies and corporate 
% sponsors that provided financial support.

% \clearpage

\section*{Impact Statement}

This paper presents work whose goal is to advance the field of machine learning. There are many potential societal consequences of our work, none of which we feel must be specifically highlighted here.


\bibliography{example_paper}

\begin{thebibliography}{36}
\providecommand{\natexlab}[1]{#1}
\providecommand{\url}[1]{\texttt{#1}}
\expandafter\ifx\csname urlstyle\endcsname\relax
  \providecommand{\doi}[1]{doi: #1}\else
  \providecommand{\doi}{doi: \begingroup \urlstyle{rm}\Url}\fi

\bibitem[Balanya et~al.(2022)Balanya, Maro{\~n}as, and Ramos]{balanya2022adaptive}
Balanya, S.~A., Maro{\~n}as, J., and Ramos, D.
\newblock Adaptive temperature scaling for robust calibration of deep neural networks.
\newblock \emph{arXiv preprint arXiv:2208.00461}, 2022.

\bibitem[Barbu et~al.(2019)Barbu, Mayo, Alverio, Luo, Wang, Gutfreund, Tenenbaum, and Katz]{barbu2019objectnet}
Barbu, A., Mayo, D., Alverio, J., Luo, W., Wang, C., Gutfreund, D., Tenenbaum, J., and Katz, B.
\newblock Objectnet: A large-scale bias-controlled dataset for pushing the limits of object recognition models.
\newblock \emph{Advances in neural information processing systems}, 32, 2019.

\bibitem[Beery et~al.(2020)Beery, Cole, and Gjoka]{beery2020iwildcam}
Beery, S., Cole, E., and Gjoka, A.
\newblock The iwildcam 2020 competition dataset.
\newblock \emph{arXiv preprint arXiv:2004.10340}, 2020.

\bibitem[Corbiere et~al.(2021)Corbiere, Thome, Saporta, Vu, Cord, and Perez]{corbiere2021confidence}
Corbiere, C., Thome, N., Saporta, A., Vu, T.-H., Cord, M., and Perez, P.
\newblock Confidence estimation via auxiliary models.
\newblock \emph{IEEE Transactions on Pattern Analysis and Machine Intelligence}, 44\penalty0 (10):\penalty0 6043--6055, 2021.

\bibitem[Darlow et~al.(2018)Darlow, Crowley, Antoniou, and Storkey]{darlow2018cinic}
Darlow, L.~N., Crowley, E.~J., Antoniou, A., and Storkey, A.~J.
\newblock Cinic-10 is not imagenet or cifar-10.
\newblock \emph{arXiv preprint arXiv:1810.03505}, 2018.

\bibitem[Deng et~al.(2009)Deng, Dong, Socher, Li, Li, and Fei-Fei]{deng2009imagenet}
Deng, J., Dong, W., Socher, R., Li, L.-J., Li, K., and Fei-Fei, L.
\newblock Imagenet: A large-scale hierarchical image database.
\newblock In \emph{2009 IEEE conference on computer vision and pattern recognition}, pp.\  248--255. Ieee, 2009.

\bibitem[Deng et~al.(2022)Deng, Gould, and Zheng]{deng2022strong}
Deng, W., Gould, S., and Zheng, L.
\newblock On the strong correlation between model invariance and generalization.
\newblock In \emph{Advances in Neural Information Processing Systems}, 2022.

\bibitem[Dirac(1981)]{dirac1981principles}
Dirac, P. A.~M.
\newblock \emph{The principles of quantum mechanics}.
\newblock Number~27. Oxford university press, 1981.

\bibitem[Gong et~al.(2021)Gong, Lin, Yao, Dietterich, Divakaran, and Gervasio]{gong2021confidence}
Gong, Y., Lin, X., Yao, Y., Dietterich, T.~G., Divakaran, A., and Gervasio, M.
\newblock Confidence calibration for domain generalization under covariate shift.
\newblock In \emph{Proceedings of the IEEE/CVF International Conference on Computer Vision}, pp.\  8958--8967, 2021.

\bibitem[Guo et~al.(2017)Guo, Pleiss, Sun, and Weinberger]{guo2017calibration}
Guo, C., Pleiss, G., Sun, Y., and Weinberger, K.~Q.
\newblock On calibration of modern neural networks.
\newblock In \emph{International conference on machine learning}, pp.\  1321--1330. PMLR, 2017.

\bibitem[Gupta et~al.(2020)Gupta, Rahimi, Ajanthan, Mensink, Sminchisescu, and Hartley]{gupta2020calibration}
Gupta, K., Rahimi, A., Ajanthan, T., Mensink, T., Sminchisescu, C., and Hartley, R.
\newblock Calibration of neural networks using splines.
\newblock \emph{arXiv preprint arXiv:2006.12800}, 2020.

\bibitem[Hebbalaguppe et~al.(2022)Hebbalaguppe, Prakash, Madan, and Arora]{hebbalaguppe2022stitch}
Hebbalaguppe, R., Prakash, J., Madan, N., and Arora, C.
\newblock A stitch in time saves nine: A train-time regularizing loss for improved neural network calibration.
\newblock In \emph{Proceedings of the IEEE/CVF Conference on Computer Vision and Pattern Recognition}, pp.\  16081--16090, 2022.

\bibitem[Hendrycks \& Dietterich(2019)Hendrycks and Dietterich]{hendrycks2019benchmarking}
Hendrycks, D. and Dietterich, T.
\newblock Benchmarking neural network robustness to common corruptions and perturbations.
\newblock \emph{arXiv preprint arXiv:1903.12261}, 2019.

\bibitem[Hendrycks et~al.(2021{\natexlab{a}})Hendrycks, Basart, Mu, Kadavath, Wang, Dorundo, Desai, Zhu, Parajuli, Guo, Song, Steinhardt, and Gilmer]{hendrycks2021many}
Hendrycks, D., Basart, S., Mu, N., Kadavath, S., Wang, F., Dorundo, E., Desai, R., Zhu, T., Parajuli, S., Guo, M., Song, D., Steinhardt, J., and Gilmer, J.
\newblock The many faces of robustness: A critical analysis of out-of-distribution generalization.
\newblock \emph{ICCV}, 2021{\natexlab{a}}.

\bibitem[Hendrycks et~al.(2021{\natexlab{b}})Hendrycks, Zhao, Basart, Steinhardt, and Song]{hendrycks2021nae}
Hendrycks, D., Zhao, K., Basart, S., Steinhardt, J., and Song, D.
\newblock Natural adversarial examples.
\newblock \emph{CVPR}, 2021{\natexlab{b}}.

\bibitem[Joy et~al.(2023)Joy, Pinto, Lim, Torr, and Dokania]{joy2023sample}
Joy, T., Pinto, F., Lim, S.-N., Torr, P.~H., and Dokania, P.~K.
\newblock Sample-dependent adaptive temperature scaling for improved calibration.
\newblock In \emph{Proceedings of the AAAI Conference on Artificial Intelligence}, volume~37, pp.\  14919--14926, 2023.

\bibitem[Jung et~al.(2023)Jung, Seo, Jeong, and Choi]{jung2023scaling}
Jung, S., Seo, S., Jeong, Y., and Choi, J.
\newblock Scaling of class-wise training losses for post-hoc calibration.
\newblock \emph{ICML}, 2023.

\bibitem[Krishnan \& Tickoo(2020)Krishnan and Tickoo]{krishnan2020improving}
Krishnan, R. and Tickoo, O.
\newblock Improving model calibration with accuracy versus uncertainty optimization.
\newblock \emph{Advances in Neural Information Processing Systems}, 33:\penalty0 18237--18248, 2020.

\bibitem[Krizhevsky et~al.(2009)Krizhevsky, Hinton, et~al.]{krizhevsky2009learning}
Krizhevsky, A., Hinton, G., et~al.
\newblock Learning multiple layers of features from tiny images.
\newblock 2009.

\bibitem[Kull et~al.(2019)Kull, Perello~Nieto, K{\"a}ngsepp, Silva~Filho, Song, and Flach]{kull2019beyond}
Kull, M., Perello~Nieto, M., K{\"a}ngsepp, M., Silva~Filho, T., Song, H., and Flach, P.
\newblock Beyond temperature scaling: Obtaining well-calibrated multi-class probabilities with dirichlet calibration.
\newblock \emph{Advances in neural information processing systems}, 32, 2019.

\bibitem[Kumar et~al.(2019)Kumar, Liang, and Ma]{kumar2019verified}
Kumar, A., Liang, P.~S., and Ma, T.
\newblock Verified uncertainty calibration.
\newblock \emph{Advances in Neural Information Processing Systems}, 32, 2019.

\bibitem[Liang et~al.(2020)Liang, Zhang, Wang, and Jacobs]{liang2020improved}
Liang, G., Zhang, Y., Wang, X., and Jacobs, N.
\newblock Improved trainable calibration method for neural networks on medical imaging classification.
\newblock \emph{arXiv preprint arXiv:2009.04057}, 2020.

\bibitem[Mukhoti et~al.(2020)Mukhoti, Kulharia, Sanyal, Golodetz, Torr, and Dokania]{mukhoti2020calibrating}
Mukhoti, J., Kulharia, V., Sanyal, A., Golodetz, S., Torr, P., and Dokania, P.
\newblock Calibrating deep neural networks using focal loss.
\newblock \emph{Advances in Neural Information Processing Systems}, 33:\penalty0 15288--15299, 2020.

\bibitem[Nixon et~al.(2019)Nixon, Dusenberry, Zhang, Jerfel, and Tran]{nixon2019measuring}
Nixon, J., Dusenberry, M.~W., Zhang, L., Jerfel, G., and Tran, D.
\newblock Measuring calibration in deep learning.
\newblock In \emph{CVPR workshops}, volume~2, 2019.

\bibitem[Tao et~al.(2023)Tao, Dong, and Xu]{tao2023dual}
Tao, L., Dong, M., and Xu, C.
\newblock Dual focal loss for calibration.
\newblock \emph{ICML}, 2023.

\bibitem[Tomani et~al.(2021)Tomani, Gruber, Erdem, Cremers, and Buettner]{tomani2021post}
Tomani, C., Gruber, S., Erdem, M.~E., Cremers, D., and Buettner, F.
\newblock Post-hoc uncertainty calibration for domain drift scenarios.
\newblock In \emph{Proceedings of the IEEE/CVF Conference on Computer Vision and Pattern Recognition}, pp.\  10124--10132, 2021.

\bibitem[Tomani et~al.(2022)Tomani, Cremers, and Buettner]{tomani2022parameterized}
Tomani, C., Cremers, D., and Buettner, F.
\newblock Parameterized temperature scaling for boosting the expressive power in post-hoc uncertainty calibration.
\newblock In \emph{European Conference on Computer Vision}, pp.\  555--569. Springer, 2022.

\bibitem[Tomani et~al.(2023)Tomani, Waseda, Shen, and Cremers]{tomani2023beyond}
Tomani, C., Waseda, F.~K., Shen, Y., and Cremers, D.
\newblock Beyond in-domain scenarios: robust density-aware calibration.
\newblock In \emph{International Conference on Machine Learning}, pp.\  34344--34368. PMLR, 2023.

\bibitem[Wang et~al.(2019)Wang, Ge, Lipton, and Xing]{wang2019learning}
Wang, H., Ge, S., Lipton, Z., and Xing, E.~P.
\newblock Learning robust global representations by penalizing local predictive power.
\newblock In \emph{Advances in Neural Information Processing Systems}, pp.\  10506--10518, 2019.

\bibitem[Wang et~al.(2023)Wang, Koniusz, Gedeon, and Zheng]{wang2023adaptive}
Wang, L., Koniusz, P., Gedeon, T., and Zheng, L.
\newblock Adaptive multi-head contrastive learning.
\newblock \emph{arXiv preprint arXiv:2310.05615}, 2023.

\bibitem[Wang et~al.(2020)Wang, Long, Wang, and Jordan]{wang2020transferable}
Wang, X., Long, M., Wang, J., and Jordan, M.
\newblock Transferable calibration with lower bias and variance in domain adaptation.
\newblock \emph{Advances in Neural Information Processing Systems}, 33:\penalty0 19212--19223, 2020.

\bibitem[Wightman(2019)]{rw2019timm}
Wightman, R.
\newblock Pytorch image models.
\newblock \url{https://github.com/rwightman/pytorch-image-models}, 2019.

\bibitem[Xia \& Bouganis(2023)Xia and Bouganis]{xia2023window}
Xia, G. and Bouganis, C.-S.
\newblock Window-based early-exit cascades for uncertainty estimation: When deep ensembles are more efficient than single models.
\newblock \emph{arXiv preprint arXiv:2303.08010}, 2023.

\bibitem[Xiong et~al.(2024)Xiong, Deng, Koh, Wu, Li, Xu, and Hooi]{xiong2024proximity}
Xiong, M., Deng, A., Koh, P. W.~W., Wu, J., Li, S., Xu, J., and Hooi, B.
\newblock Proximity-informed calibration for deep neural networks.
\newblock \emph{Advances in Neural Information Processing Systems}, 36, 2024.

\bibitem[Zhang et~al.(2020)Zhang, Kailkhura, and Han]{zhang2020mix}
Zhang, J., Kailkhura, B., and Han, T. Y.-J.
\newblock Mix-n-match: Ensemble and compositional methods for uncertainty calibration in deep learning.
\newblock In \emph{International conference on machine learning}, pp.\  11117--11128. PMLR, 2020.

\bibitem[Zou et~al.(2023)Zou, Deng, and Zheng]{zou2023adaptive}
Zou, Y., Deng, W., and Zheng, L.
\newblock Adaptive calibrator ensemble: Navigating test set difficulty in out-of-distribution scenarios.
\newblock In \emph{Proceedings of the IEEE/CVF International Conference on Computer Vision}, pp.\  19333--19342, 2023.

\end{thebibliography}
\bibliographystyle{icml2025}

%%%%%%%%%%%%%%%%%%%%%%%%%%%%%%%%%%%%%%%%%%%%%%%%%%%%%%%%%%%%%%%%%%%%%%%%%%%%%%%
%%%%%%%%%%%%%%%%%%%%%%%%%%%%%%%%%%%%%%%%%%%%%%%%%%%%%%%%%%%%%%%%%%%%%%%%%%%%%%%
% APPENDIX
%%%%%%%%%%%%%%%%%%%%%%%%%%%%%%%%%%%%%%%%%%%%%%%%%%%%%%%%%%%%%%%%%%%%%%%%%%%%%%%
%%%%%%%%%%%%%%%%%%%%%%%%%%%%%%%%%%%%%%%%%%%%%%%%%%%%%%%%%%%%%%%%%%%%%%%%%%%%%%%
\newpage
\appendix
\onecolumn

\section{Our Algorithm}
\label{suppl:alg}

Alg.~\ref{algorithm_1} shows the calibrator with our proposed Correctness-Aware loss.

\begin{algorithm}
\caption{Calibrator with our proposed Correctness-Aware Loss}
\label{algorithm_1}
\begin{algorithmic}
\footnotesize
\STATE{\textbf{Input}: a classification model $f$ to be calibrated ($f'(\cdot)$ extracts the logit vector, and $\sigma(\cdot)$ denotes the softmax function), a calibrator $g$ parameterized by $\theta$, the total number of transforms $M$, and $k$ for selecting the top-$k$ maximum softmax scores. $y$ and $\hat{y}$ denote the ground truth label and the predicted label for a given image sample $\mX$, respectively.}
\STATE{\textbf{Step 1:} Obtain the logit vector: $\vz=f'(\mX)$ and the softmax vector: $\vv=\sigma(\vz)$.}
\STATE{\textbf{Step 2:} Apply $M$ transforms to the original input image $\mX$ to obtain its transformed images $\mX^{(i)}$ ($i\in\idx{M}$), then obtain their corresponding softmax vectors: $\vv_i=\sigma(f'(\mX^{(i)}))$.} 
\STATE{\textbf{Step 3:} Get the indices $\vq$ of the top-$k$ maximum softmax scores from the logit vector $\vv$ using Eq.~\eqref{eq:topkk}.}
\STATE{\textbf{Step 4:} Use the generated indices $\vq$ to form new vectors $\vv_i \in \mbr{k}$, concatenate these new $k$-dimensional vectors resulting in $\oplus_{i\in\idx{M}}\vv_i[\mathbf{q}] \in \mbr{M \times k}$, then pass this resulting matrix to the calibrator $g$ to produce the temperature $\tau$ via Eq.~\eqref{eq:concate}.}
\STATE{\textbf{Step 5:} Apply the learned temperature $\tau$ to the original logit vector and obtain its maximum softmax score via Eq.~\eqref{eq:max}.}
\STATE{\textbf{Step 6:} Plug the updated maximum softmax prediction score from Step 5, the ground truth label $y$, and the predicted label $\hat{y}$ into our proposed Correctness-Aware Loss via Eq.~\eqref{eq:optim_goal}.}
\STATE{\textbf{Return:} Calibrator model weights $\theta$.}
% \STATE{\textbf{Input}: a classification model $f$ to be calibrated, a calibrator $g$ parameterized by $\theta$, a given image sample $\mX$, the total number of transforms $M$, and $k$ for selecting the top-$k$ maximum softmax scores. Softmax function $\sigma(\cdot)$.}
% \STATE{\textbf{Step 1:} Obtaining logit vector: $\vz=f(\mX)$} and softmax vector: $\vv=\sigma(\vz)$.
% \STATE{\textbf{Step 2:} Applying $M$ transforms to original input image $\mX$ to get its transformed images $\mX^{(i)}$ ( $i\in\idx{M}$), and then obtaining their corresponding softmax vectors: $\vv_i=\sigma(f(\mX^{(i)}))$.} 
% \STATE{\textbf{Step 3:} Getting the top-$k$ maximum softmax score indices $\vq \in \mbr{k}$ of the logit vector $\vv$ via Eq.~\eqref{eq:topkk}.}
% \STATE{\textbf{Step 4:} Using generated indices $\vq$ to form new vectors of $\vv_i \in \mbr{k}$ ($i\in\idx{M}$), concatenating these new $k$-dimensional vectors which results in $\oplus_{i\in\idx{M}}\vv_i[\mathbf{q}] \in \mbr{M \times k}$, and then passing this resulting matrix to the calibrator $g$ to produce the temperature $\tau$ via Eq.~\eqref{eq:concate}.}
% \STATE{\textbf{Step 5:} Applying the learned temperature $\tau$ to the original logit vector and get its maximum softmax score via Eq.~\eqref{eq:max}.}
% \STATE{\textbf{Step 6:} Plugging the updated maximum softmax prediction score from step 5, ground truth label as well as the predicted label into our proposed Correctness-Aware Loss via Eq.~\eqref{eq:optim_goal}.}
\end{algorithmic}

\end{algorithm}

\section{Comparison between CA loss and MLE}\label{sec:theory_comparison}
Maximum Likelihood Estimate (\lei{MLE}) is widely used for calibration training~\citep{kumar2019verified}, under concrete formats such as the Cross-Entropy (CE) or Mean Square Error (MSE) losses. This section goes through their connections and differences with the CA loss. Appendix~\ref{sec:discussion} shows more discussions.

\textbf{For correct predictions, MLE (\eg, CE or MSE loss) has similar effect with the CA loss}. MLE enforces the softmax probability of the ground-truth class to be close to 1. For correct predictions, the softmax probability of the ground-truth class equals the sample confidence (maximum probability in the softmax vector). Under this scenario, MLE aligns with both the calibration objective and the CA loss: the confidence of correct predictions should be possibly high. 

\textbf{For wrong predictions, MLE sometimes deviates from the calibration goal while CA is theoretically consistent.} In Fig. \ref{fig:loss_surface}, we visualize the loss surface of the CE, MSE, and CA loss \wrt temperature and softmax probability on the ground truth class (first row), from which we use examples of two typical calibration training samples for more intuitive illustration (both first and second rows). Particularly, optimal temperature ($x$-axis) is achieved when the respective loss ($y$ axis) is minimum (the second row is easier to read).

For \underline{\textit{an absolutely wrong sample}} (green curves in both rows), whose probability of the wrongly predicted class is far greater than that on the ground-truth class, the optimization direction of MLE is similar to CA: the loss curve keeps decreasing and finally a large temperature or a low confidence is obtained. In fact, under this scenario, calibration objective requires the probability of the ground-truth class to increase and probability of the wrongly predicted class to decrease. This is consistent with the objective of MLE: to increase probability on the ground-truth class.

For \underline{\textit{a narrowly wrong sample}} (yellow curves in both rows), whose probability on the wrongly predicted class is much closer to that on the ground-truth class, the optimization direction of MLE is very different from or even opposite to CA. Take the yellow curves in the second row of Fig. \ref{fig:loss_surface} as example. The CE loss, to become smaller, leads to a small temperature, meaning a large confidence, which is undesirable for this wrongly predicted sample. For MSE, its minimum is achieved when temperature is around 1.0, which does not change the temperature and confidence much. %is a relatively small value and means a relatively large confidence. 
This again is undesirable. In comparison, the CA loss keep decreasing when temperature increases so will eventually give a large temperature or a small confidence for this type of samples. This is consistent with the calibration objective. 

Empirically, we find that such narrowly wrong predictions take up 2\%-8\% of the calibration set (ImageNet validation).\footnote{We first compute the ratio of the probability on the ground-truth class to that on the wrongly predicted class. We define a sample is narrowly wrong prediction if this ratio is higher than 0.5.} This would negatively impact training efficacy of MLE. Moreover, during inference, if a test set has many such narrowly wrong predictions, MLE will also be negatively impacted because of its unsuitable in dealing with such samples during training. This would explain why our system is superior to and on par with state of the art on OOD and IND test sets, respectively (refer to Sec.~\ref{sec:exp}). In both IND and OOD scenarios, our calibrated models are much better.% than uncalibrated ones.

\section{Compared to Pre-hoc Calibration Methods}
Although our method focuses on post-hoc model calibration, we compare our method with pre-hoc calibration methods here.
We keep the calibrator networks and inputs the same, replacing only the CA loss with the losses from MDCA \cite{hebbalaguppe2022stitch} and DCA \cite{liang2020improved}. The results are shown in Table~\ref{tab:pre_hoc}. We observe that our CA loss outperforms DCA and MDCA.

We also explore potential complementary strengths between train-time calibration methods and our approach. To this end, we combined our CA loss with the loss in MDCA or DCA to train an image transformation-based calibrator. However, we cannot observe complementary strengths of MDCA loss or DCA loss to CA loss in Fig~\ref{tab:pre_hoc}. This suggests that some training-time calibration methods may not directly benefit the post-hoc calibration system. They may also face the "narrowly wrong prediction issue". 

\begin{table*}[t]
\centering
\setlength{\tabcolsep}{1.5mm}{
    
    \caption{\textbf{Calibrator comparison under the ImageNet setup.} Each reported number is averaged over 10 classifiers. We use four test sets: ImageNet-A, ImageNet-R, ImageNet-S, and ObjectNet, and four metrics: ECE (bin=25), BS, KS, and AUC (AUROC). Best results in each column are in bold.}
    \vspace{-0.2cm}
    \label{tab:pre_hoc}
\scriptsize
    \begin{tabular}{l cccc cccc cccc cccc}
            \toprule
            \multirow{2}{*}{\textbf{Method}} & \multicolumn{4}{c}{ImageNet-A} & \multicolumn{4}{c}{ImageNet-R} & \multicolumn{4}{c}{ImageNet-S} & \multicolumn{4}{c}{ObjectNet} \\
            \cmidrule(lr){2-5} \cmidrule(lr){6-9} \cmidrule(lr){10-13} \cmidrule(lr){14-17} 
            & ECE$\downarrow$ & BS$\downarrow$ & KS$\downarrow$ & AUC$\uparrow$ & ECE$\downarrow$ & BS$\downarrow$ & KS$\downarrow$ & AUC$\uparrow$ & ECE$\downarrow$ & BS$\downarrow$ & KS$\downarrow$ & AUC$\uparrow$ & ECE$\downarrow$ & BS$\downarrow$ & KS$\downarrow$ & AUC$\uparrow$\\
            \midrule
            Uncal & 39.44 & 32.90 & 43.47 & 61.87 & 13.97 & 16.89 & 19.90 & 88.06 & 20.92 & 21.67 & 29.01 & 82.57 & 31.21 & 25.20 & 36.48 & 78.05 \\
            DCA & 32.10 & 25.32 & 36.33 & 63.33 & 10.00 & 14.03 & 17.99 & 89.05 & 17.47 & 18.00 & 25.07 & 84.04 & 25.25 & 20.75 & 31.13 & 78.54 \\
            MDCA & 27.80 & 21.71 & 32.24 & 63.32 & 7.37 & 13.85 & 16.28 & 88.04 & 13.09 & 16.97 & 22.51 & 82.70 & 21.11 & 18.49 & 28.02 & 78.05 \\
            CA (ours) & \textbf{20.65} & \textbf{16.79} & \textbf{22.50} & \textbf{63.74} & \textbf{4.91} & \textbf{12.21} & \textbf{10.12} & \textbf{90.22} & \textbf{4.00} & \textbf{13.83} & \textbf{13.12} & \textbf{84.87} & \textbf{10.33} & \textbf{14.59} & \textbf{18.72} & \textbf{79.25} \\
            DCA+CA & 30.31 & 24.00 & 34.56 & 63.45 & 8.74 & 13.50 & 16.90 & 89.25 & 15.29 & 17.01 & 23.34 & 84.32 & 23.32 & 19.66 & 29.51 & 78.65 \\
            MDCA+CA & 26.36 & 20.61 & 30.63 & 63.52 & 6.59 & 13.41 & 15.31 & 88.38 & 11.18 & 16.20 & 21.04 & 83.10 & 19.38 & 17.61 & 26.57 & 78.22 \\
            \bottomrule
    \end{tabular}
    }
    \vspace{-0.2cm}
\end{table*}

\section{Further Discussion}
\label{sec:discussion}

\textbf{How could the CA loss improve the ECE metric?} ECE  bins confidence and calculates the difference between confidence and accuracy of samples in each bin. In the extreme case where bin size is infinitely small, each bin will contain only one sample (assuming no image duplicates), meaning accuracy of each bin is either 100\% or 0\%. In this scenario, the CA loss will push correct (wrong) predictions to the high (low) confidence, which always reduces ECE. When bin size gradually becomes larger, improvement brought by CA loss will be less definite but still visible.

\textbf{Sample-adaptive temperature} used in~\cite{joy2023sample,balanya2022adaptive,wang2023adaptive} and our method has different properties from global temperature~\cite{guo2017calibration}. % While many existing methods train and use a global temperature for calibration, our method and a few recent method \blue{PTS, PROCAL} predict different temperatures for different samples. 
Because global temperature does not change the order of samples ranked by their confidence, it cannot improve the ability of confidence to separate correct and wrong predictions. Sample-adaptive temperature at least has such potential (but under specific design). % comparison, sample-adaptive temperatu which potentially is very beneficial for users during decision making.  sample separability 
On the other hand, trained with the CE loss, the sample-adaptive temperature is demonstrated to produce a competitive calibrator for IND test sets~\cite{joy2023sample,balanya2022adaptive}. But issues with CE and the lack of using additional information limit its effectiveness for OOD data. In comparison, our method is competitive on both IND and OOD  test sets. %because of the CA loss and use of transformed images as calibrator input. 

\begin{wrapfigure}{tl}{0.49\textwidth} 
\vspace{-0.50cm}
\centering
\includegraphics[width=0.49\textwidth]{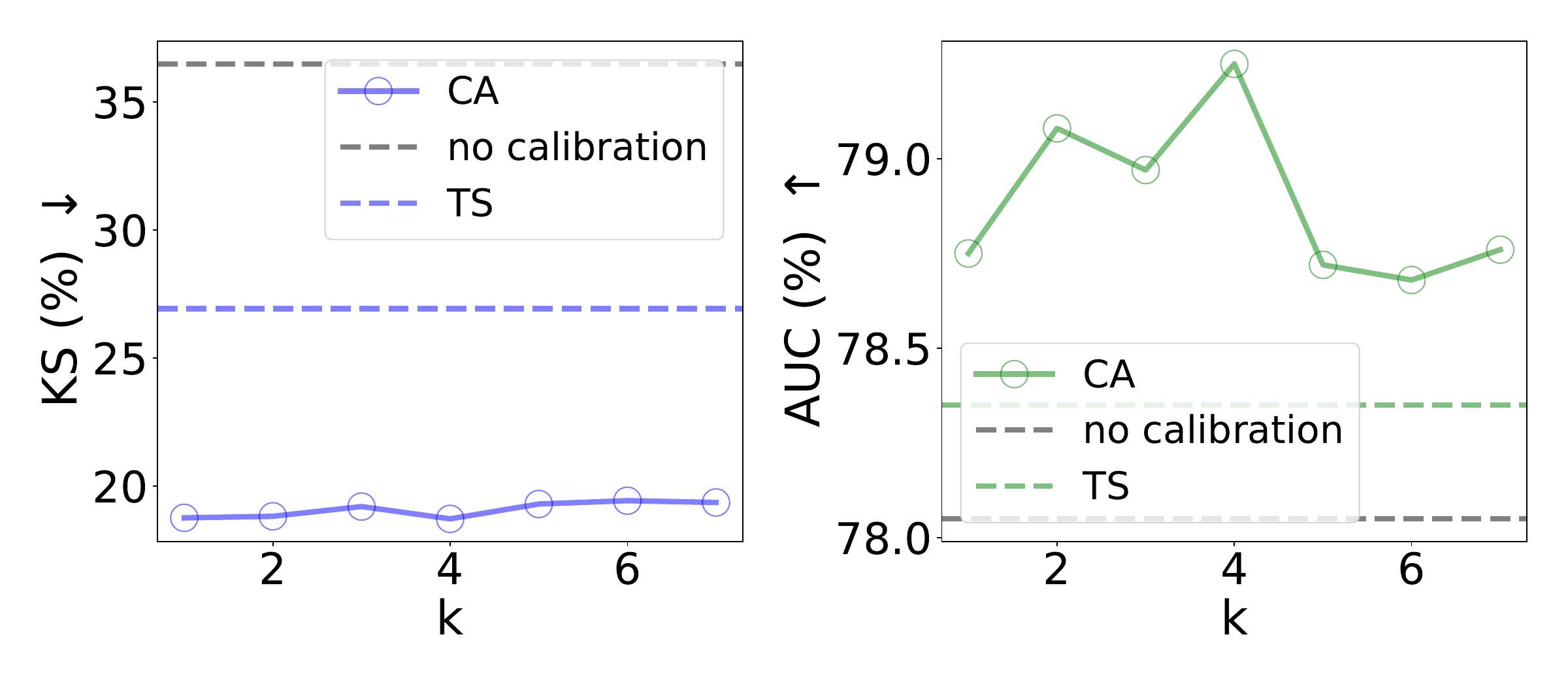}
\vspace{-8mm}
\caption{\textbf{Impact of $k$ in top-$k$ index selection (Sec. \ref{Sec:network}).} % of variants of our method with different }. 
%We report KS (\%) and AUROC (\%) 
We use ObjectNet test set. Under various $k$ our method is better (lower KS and higher AUROC) than uncalibrated models and TS. We choose $k\!=\!4$ as trade-off between performance and computational cost. 
}  
\label{fig:topk}
\vspace{-0.50cm}
\end{wrapfigure}

\textbf{\lei{Why the CA loss sometimes still have empirical failures?}} A calibrator perfectly optimized by the CA loss will give 0 ECE, because all the correctly (wrongly) classified samples will have confidence of 1 (0). In practice, however, the bottleneck is to tell prediction correctness. We use augmented images but it might not be an optimal solution. In future we will explore new methods for correctness prediction. % as future work, to improve this method. 

\textbf{Correctness prediction performance as a potential calibration metric.} Given a confidence value, better separability between correct and wrong predictions leads to safer decision making for users, because less mistakes are made. This paper uses area under the ROC curve (AUROC) to measure the performance of predicting classification correctness. As shown in Sec.~\ref{appendix-auroc}, using a calibration method does not always mean user makes less mistakes during decision making. Considering the strong tie between this 2-way classification problem and model calibration (Eq.~\eqref{eq:goal}), we think AUROC can be an additional evaluation metric for model calibration. 
% In Sec.~\ref{appendix-auroc}, more comparisons on AUROC between various methods are presented. 

\textbf{Impact of $k$.} We use the indices of top-$k$ confidences to locate and select the $k$-dim Softmax vectors from the transformed images. In Fig.~\ref{fig:topk}, we find that for various values of $k$ our method improves over uncalibrated models. Moreover, $k\!>\!5$ does not bring much improvement. Considering computational cost, we use $k\!=\!4$. Note $k$ is chosen on the ImageNet-A test set and applied on all the other test sets.  %the best calibration performance is achieved when $k=3$. We choose 

\textbf{Computational cost.} On a server with 1 GeForce RTX 3090 GPU, it takes our method 583 seconds to train a calibrator on ImageNet-Val;
in comparison, it take PTS and temperature scaling 987 seconds and 32 seconds respectively in training. Because temperature scaling only learns a single parameter (\ie, temperature), it is the quickest to train. The inference time for ours, temperature scaling, and PTS is similar: 2.33, 1.63, and 2.8 milliseconds per image, respectively. The time complexity of our method is the same as that of PTS.

\textbf{The significance of the bounds of CA loss}.
The significance of the bounds for the CA loss lies in understanding its behavior during training:
\begin{itemize}
    \item Lower bound: Demonstrates whether the loss function can converge during the optimization process, ensuring stability and effectiveness.
    \item Upper bound: Indicates the maximum possible value of the loss, useful for diagnosing training stability and ensuring expected behavior.
\end{itemize}

\section{Access Of Benchmarks And Models}
\label{appendix-exp}

In this section, we introduce the benchmarking datasets and classification models used in our paper.

% \subsection{ImageNet Setup}
% \label{sec:imagenet_setup}

\textbf{ImageNet models.}
We employ the ImageNet models from the PyTorch Image Models (timm) library \citep{rw2019timm}, which offers models trained or fine-tuned on the ImageNet-1k training set \citep{deng2009imagenet}. The models utilized in our paper are listed below:

\textit{\{
`beit\_base\_patch16\_384', `tv\_resnet152', `tv\_resnet50', `tv\_resnet101', `densenet121', `inception\_v4', `densenet201', `vit\_base\_patch16\_384', `deit\_base\_patch16\_224', `inception\_v3' 
\}
}

\textbf{Datasets.}
We present the test sets employed in the main paper to evaluate the aforementioned ImageNet models. Datasets mentioned below can be accessed publicly via the provided links.

\noindent\emph{ImageNet-A(dversarial)} \citep{hendrycks2021nae}: \textcolor{blue}{https://github.com/hendrycks/natural-adv-examples}. \\
\emph{ImageNet-S(ketch)} \citep{wang2019learning}: \textcolor{blue}{https://github.com/HaohanWang/ImageNet-Sketch}.\\
\emph{ImageNet-R(endition)} \citep{hendrycks2021many}: \textcolor{blue}{https://github.com/hendrycks/imagenet-r}. \\
\emph{ImageNet-Blur} \citep{hendrycks2019benchmarking}: \textcolor{blue}{https://github.com/hendrycks/robustness}.\\
% \emph{ImageNet-V2} \citep{recht2019imagenet}: \\\textcolor{blue}{https://github.com/modestyachts/ImageNetV2}.\\
\emph{ObjectNet} \citep{barbu2019objectnet}: \textcolor{blue}{https://objectnet.dev/download.html}.\\

% \subsection{CIFAR-$10$ Setup}
% \label{sec:cifar10_setup}

\textbf{CIFAR-$10$ models.}
We employ the CIFAR-$10$ models from the 
% PyTorch Image Models (timm) library \citep{rw2019timm}, 
open source library (\textcolor{blue}{https://github.com/kuangliu/pytorch-cifar})
which offers models trained or fine-tuned on the CIFAR-$10$ training set \citep{krizhevsky2009learning}. The models utilized in our paper are listed below:

\textit{\{
`VGG19', `DenseNet121', `DenseNet201', `ResNet18', `ResNet50', `ShuffleNetV2', `MobileNet', `PreActResNet101', `RegNetX\_200MF', `ResNeXt29\_2x64d'
\}
}

\textbf{Datasets.} Datasets used in the CIFAR10 setup can be found through the following links.
\noindent
\emph{CINIC} \citep{darlow2018cinic}: \\\textcolor{blue}{https://github.com/BayesWatch/cinic-10}. 
% \emph{CIFAR10} \citep{krizhevsky2009learning}: \\\textcolor{blue}{https://www.cs.toronto.edu/ kriz/cifar.html}. \\
\emph{CIFAR10-C} \cite{hendrycks2019benchmarking}(\textcolor{blue}{https://github.com/hendrycks/robustness}); \\
% \emph{CIFAR10.1} \citep{recht2018cifar10.1}: \\\textcolor{blue}{https://github.com/modestyachts/CIFAR-10.1}. \\

% \subsection{iWildCam Setup}
% \label{sec:iwildcam_setup}

\textbf{iWildCam models.}
We use the iWildCam models from the 
% PyTorch Image Models (timm) library \citep{rw2019timm}, 
open source library (\textcolor{blue}{https://worksheets.codalab.org/worksheets/0x52cea64d1d3f4fa89de326b4e31aa50a})
which offers models trained/fine-tuned on the iWildCam training set \citep{beery2020iwildcam}. The models utilized in our paper are listed below:

\textit{\{
`iwildcam\_erm\_seed1', `iwildcam\_deepCORAL\_seed0', `iwildcam\_groupDRO\_seed0', `iwildcam\_irm\_seed0',`iwildcam\_erm\_tune0', `iwildcam\_ermaugment\_tune0', `iwildcam\_ermoracle\_extraunlabeled\_tune0', `iwildcam\_swav30\_ermaugment\_seed0', `iwildcam\_dann\_coarse\_extraunlabeled\_tune0', `iwildcam\_afn\_extraunlabeled\_tune0'
\}
}

\textbf{Dataset.}
\emph{iWildCam-OOD} \citep{beery2020iwildcam} can be download from the the official guidence: \textcolor{blue}{https://github.com/p-lambda/wilds/}.

\section{Additional Visualisations}
\label{suppl:vis}

\subsection{Reliability Diagram}
\label{appendix:reliability_diagram}

\begin{figure*}[t]%
\centering
\includegraphics[width=1\textwidth]{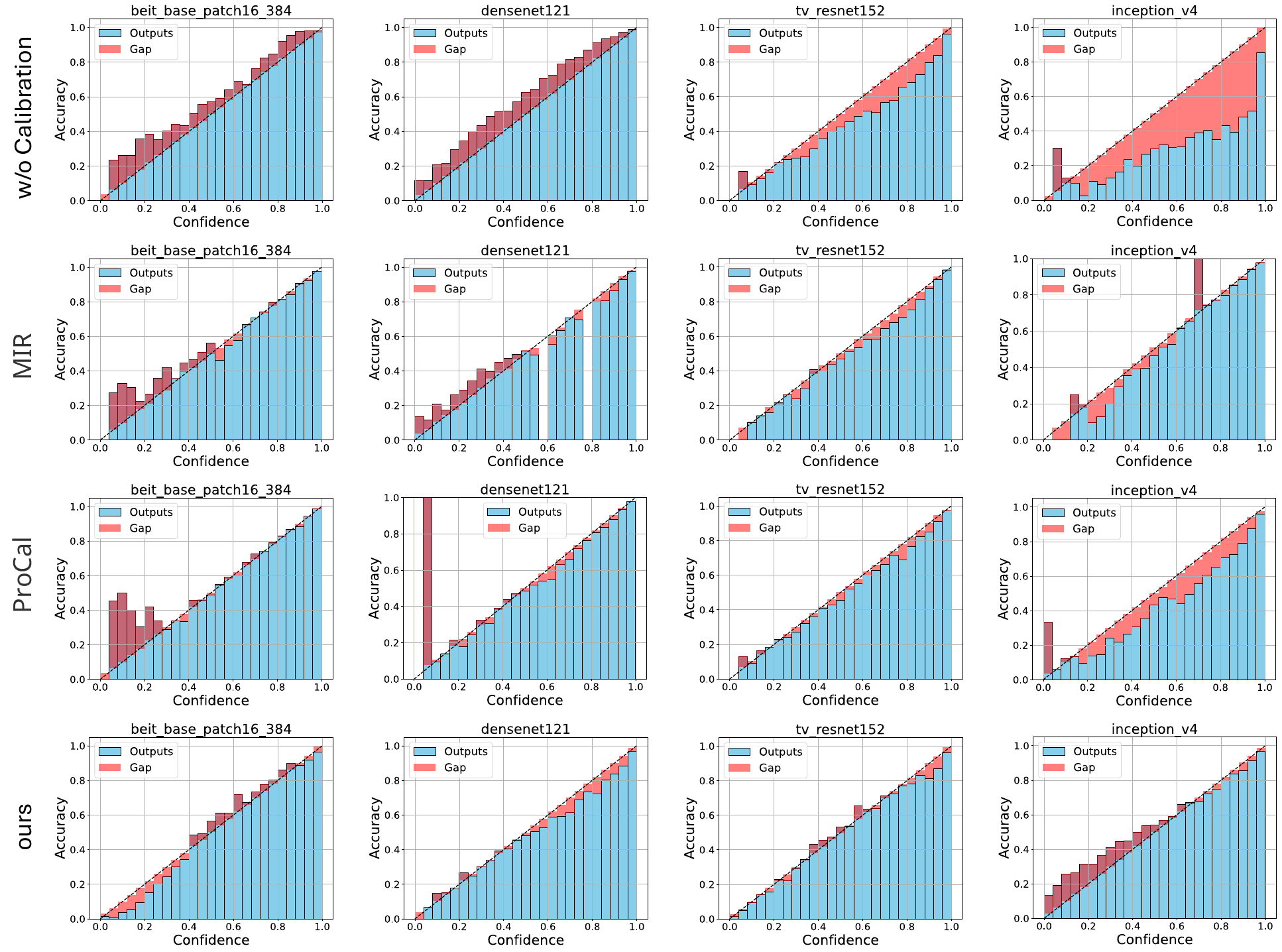}
% \vspace{-1cm}
\caption{\textbf{Reliability diagram of various models on the ImageNet validation set (25 bins)}. Bars above the dashed line indicate underconfidence, while those below indicate overconfidence. Our method effectively mitigates both overconfident and underconfident predictions across different scenarios.}
\label{fig:reliability_diagram_val}
\end{figure*}

\begin{figure*}[t]%
\centering
\includegraphics[width=1\textwidth]{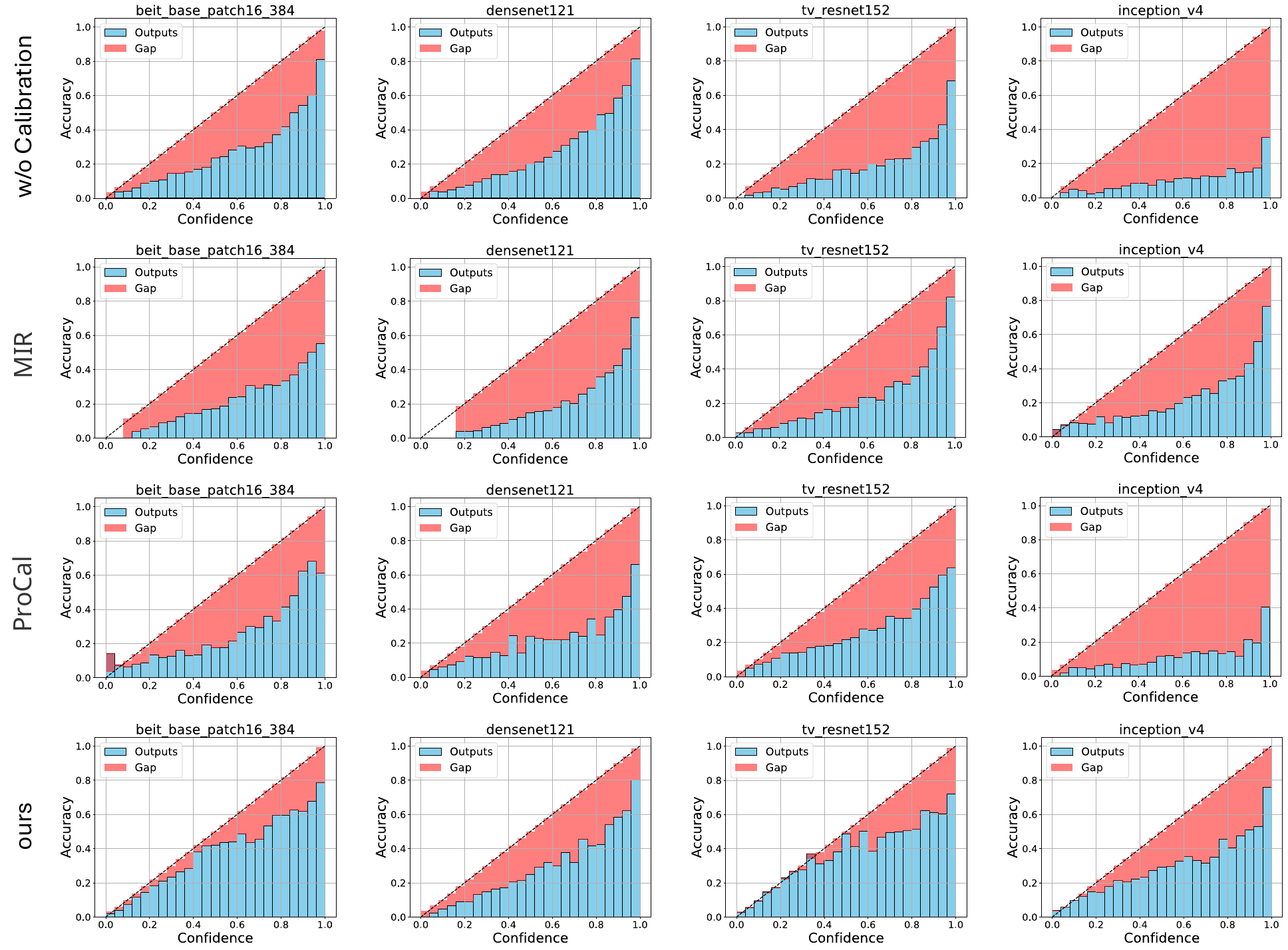}
% \vspace{-1cm}
\caption{\textbf{Reliability diagram of various models on the out-of-distribution (OOD) ObjectNet dataset (25 bins)}. Bars above the dashed line indicate underconfidence, while those below indicate overconfidence. ObjectNet is a highly challenging OOD test set, where models often exhibit severe overconfidence. Our method significantly mitigates this issue. }

\label{fig:reliability_diagram_o}
\end{figure*}

We show the reliability diagrams of various calibration methods and classification models for the in-distribution set in Fig.~\ref{fig:reliability_diagram_val} and out-of-distribution (OOD) set in Fig.~\ref{fig:reliability_diagram_o}

\subsection{AUROC curves}
\label{appendix-auroc}

\begin{figure*}[t]%
\centering
\includegraphics[width=1\textwidth]{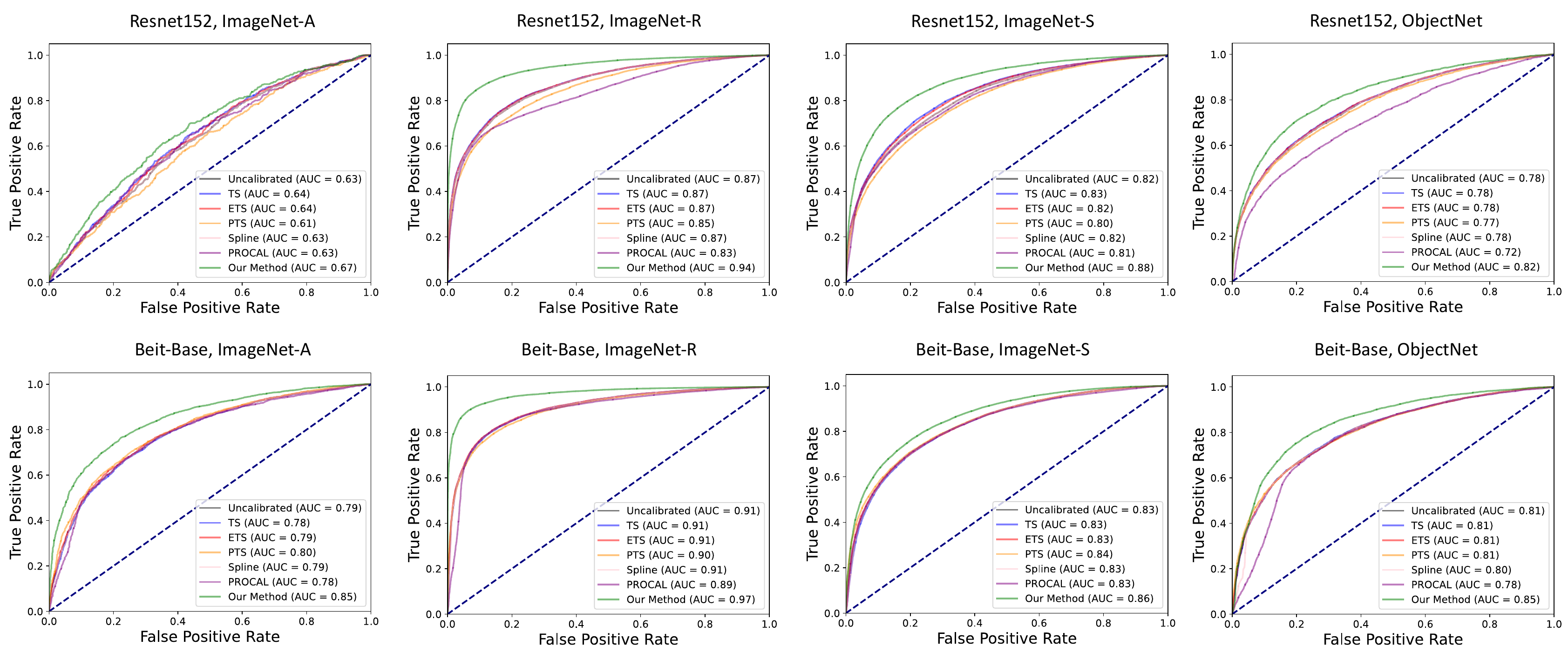}
% \vspace{-1cm}
\caption{\textbf{Comparison of Receiver Operating Characteristic (ROC) Curves Across Different Calibration Methods}. Each figure's title specifies the classifier and the test set used. It is evident that our methods (green curves) yield a higher area under ROC curve (AUROC) compared to other calibration methods, signifying an enhanced ability of our model to distinguish between correct and incorrect predictions based on calibrated confidence.
}  
\label{fig:roc_conf}
\end{figure*}

Fig.~\ref{fig:roc_conf} shows the comparison of Receiver Operating Characteristic (ROC) curves across different calibration methods.

\subsection{Distributions of predictions}
\label{appendix:more}

\begin{figure*}[tbp]
    		\centering
\subfigure[ImageNet-A]{\label{fig:imagenet-a}\includegraphics[width=0.8\textwidth]{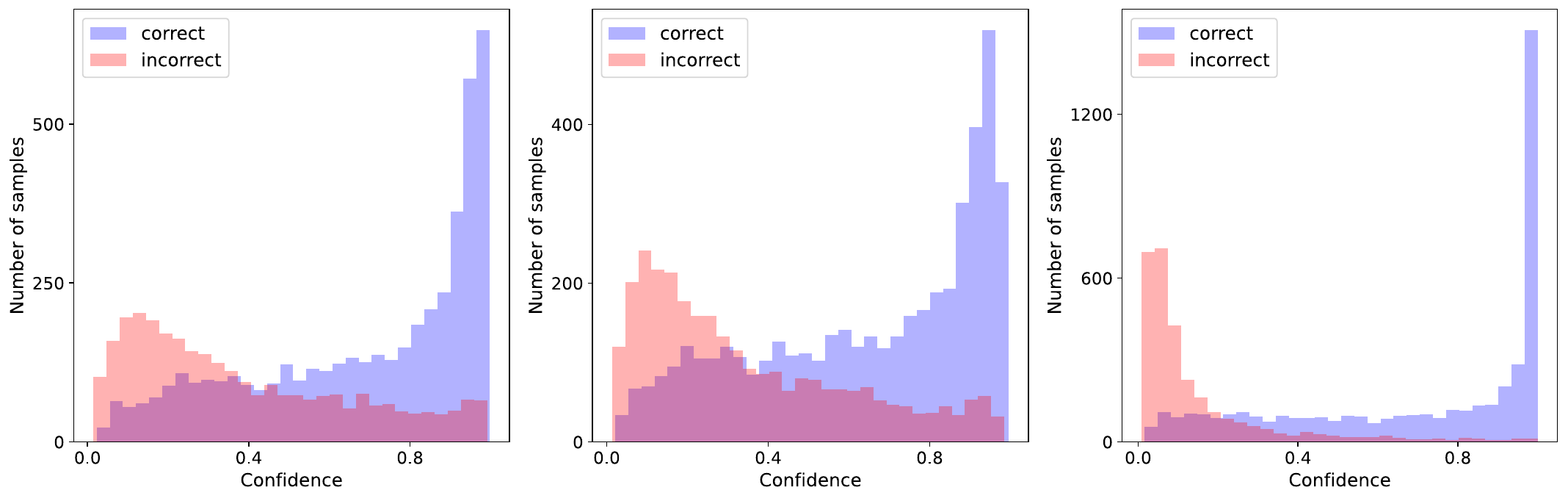}}\\
\subfigure[ImageNet-R]{\label{fig:imagenet-r}\includegraphics[width=0.8\textwidth]{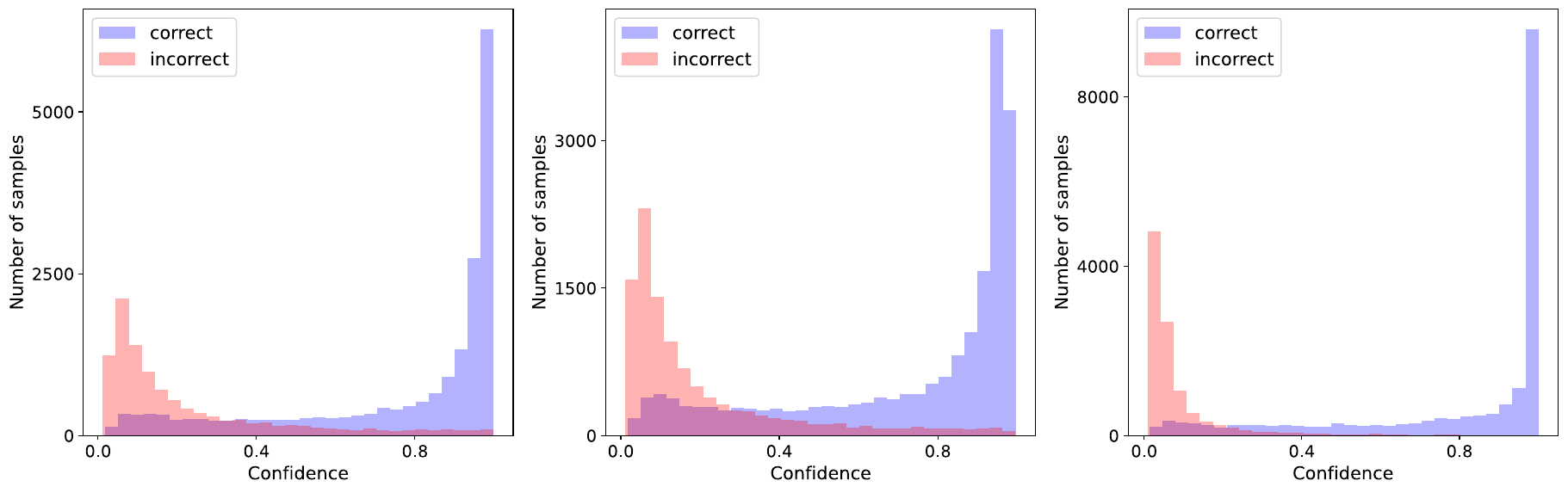}}\\
\subfigure[ImageNet-S]{\label{fig:imagenet-s}\includegraphics[width=0.8\textwidth]{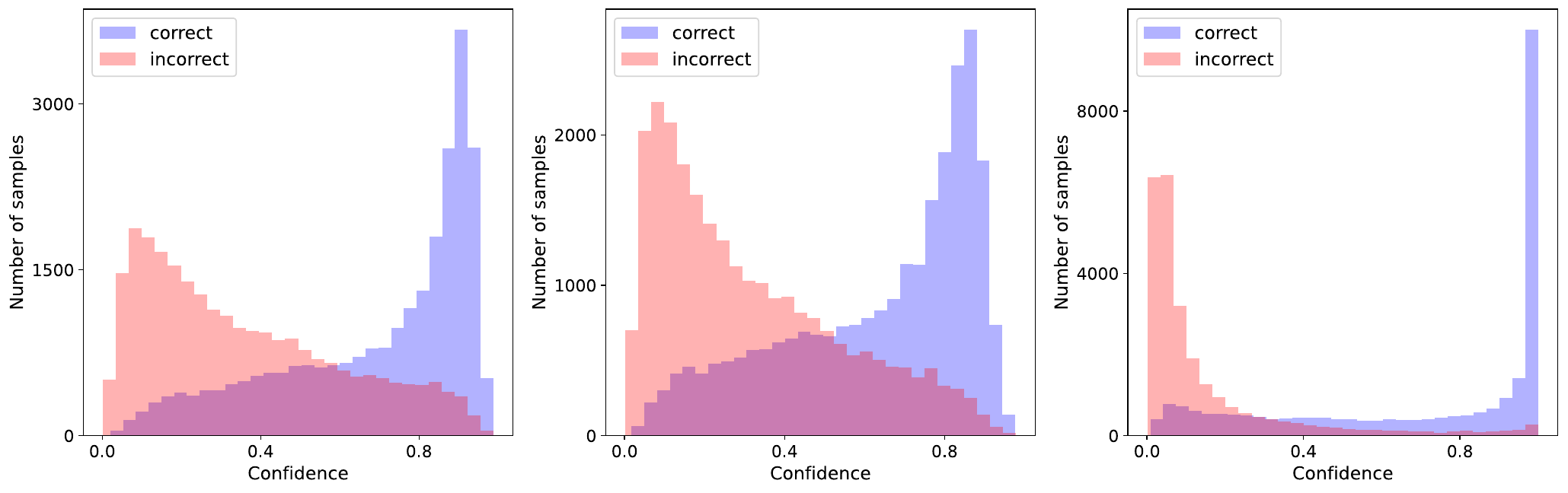}}\\
\subfigure[ObjectNet]{\label{fig:objectnet}\includegraphics[width=0.8\textwidth]{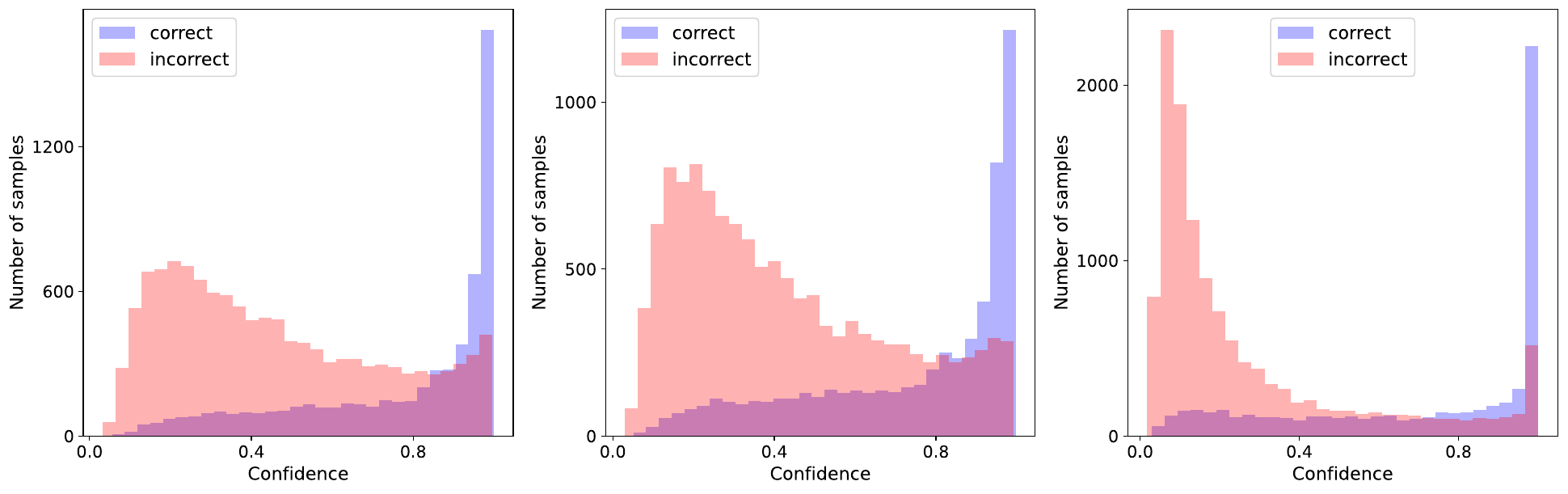}}
    %\vspace{-0.3cm}
    \captionof{figure}{Visualization of the distributions for correct and incorrect predictions of `beit\_base\_patch16\_384' on (a) ImageNet-A, (b) ImageNet-R, (c) ImageNet-S, and (d) ObjectNet. From left to right, the methods are no calibration, temperature scaling, and our method. We find that our method can better distinguish between correct and incorrect predictions by increasing the confidence value for correct predictions and decreasing it for incorrect ones.} 
  \label{fig:vis}
\end{figure*}

Visualizations of the distributions for correct and incorrect predictions on four datasets are given in Fig.~\ref{fig:vis}.

\end{document}